%% file: arxiv.tex
\documentclass{article} 
\usepackage{iclr2026_conference,times}

\input{math_commands.tex}

\usepackage{url}
\usepackage{algorithm}
\usepackage{wrapfig}
\usepackage{algpseudocode}
\usepackage{amsmath}
\usepackage{caption}
\usepackage{newfloat}
\usepackage{listings}
\usepackage{enumitem}
\usepackage{bbding}
\usepackage{amsmath}
\usepackage{amssymb}
\usepackage{mathtools}
\usepackage{amsthm}
\usepackage{pifont}
\usepackage{booktabs, multirow}
\usepackage{tabularx}
\usepackage{amssymb}
\usepackage{array}
\usepackage{titletoc}

\usepackage{makecell}

\usepackage{pifont}        
\newcommand{\cmark}{\ding{51}} 
\newcommand{\xmark}{\ding{55}} 

\usepackage{xcolor} 
\usepackage[table]{xcolor}
\definecolor{links}{HTML}{0078b0} 
\definecolor{files}{HTML}{fc6160}

\newcommand{\wy}[1]{#1}

\usepackage[
  colorlinks=true,            
  linkcolor=files,
  filecolor=files,
  urlcolor=links,
  citecolor=links
]{hyperref}

\title{ELLMob: Event-Driven Human Mobility Generation with Self-Aligned LLM Framework}

\author{%
Yusong~Wang\textsuperscript{\rm 1}\thanks{Equal contribution.}~\textnormal{,} ~Chuang~Yang\textsuperscript{\rm 2}\footnotemark[1]~\textnormal{,} ~Jiawei~Wang\textsuperscript{\rm 2}, 
~\textbf{Xiaohang Xu}\textsuperscript{\rm 2},
~\textbf{Jiayi Xu}\textsuperscript{\rm 2},
~\textbf{Dongyuan Li}\textsuperscript{\rm 2}\\
\textbf{Chuan Xiao}\textsuperscript{\rm 3}, 
~\textbf{Renhe~Jiang}\textsuperscript{\rm 2}\thanks{Corresponding author.}\\ 
\textsuperscript{\rm 1~}Institute of Science Tokyo, \textsuperscript{\rm 2~}The University of Tokyo, \textsuperscript{\rm 3~}The University of Osaka\\
\text{wangyi@lr.pi.titech.ac.jp},~\text{chuang.yang@csis.u-tokyo.ac.jp},~\text{\{jiawei,xhxu\}@g.ecc.u-tokyo.ac.jp}\\
\text{\{xujy,lidy\}@csis.u-tokyo.ac.jp},~\text{chuanx@ist.osaka-u.ac.jp},~\text{jiangrh@csis.u-tokyo.ac.jp}
}

\iclrfinalcopy 
\begin{document}

\maketitle

\begin{abstract}
Human mobility generation aims to synthesize plausible trajectory data, which is widely used in urban system research.
While Large Language Model-based methods excel at generating routine trajectories, they struggle to capture deviated mobility during large-scale societal events.
This limitation stems from two critical gaps: (1) the absence of event-annotated mobility datasets for design and evaluation, and (2) the inability of current frameworks to reconcile competitions between users' habitual patterns and event-imposed constraints when making trajectory decisions.
This work addresses these gaps with a twofold contribution.
First, we construct the first event-annotated mobility dataset covering three major events: Typhoon Hagibis, COVID-19, and the Tokyo 2021 Olympics.
Second, we propose ELLMob, a self-aligned LLM framework that first extracts competing rationales between habitual patterns and event constraints, based on Fuzzy-Trace Theory, and then iteratively aligns them to generate trajectories that are both habitually grounded and event-responsive.
Extensive experiments show that ELLMob wins state-of-the-art baselines across all events, demonstrating its effectiveness.
Our codes and datasets are available at \url{https://github.com/deepkashiwa20/ELLMob}.

\end{abstract}

\section{Introduction}

Human mobility generation aims to synthesize plausible spatio-temporal trajectories of human movement \citep{10591691}. 
The study of such trajectories offers deep insights for urban planning, transportation management, and public health \citep{Duan2023DiscoveringUM,10133883,10.1145/3637528.3671578}. 
Moreover, synthetic trajectories provide a privacy-preserving alternative that permits broader access and usage than sensitive real-world data. 
The emergence of Large Language Models (LLMs) has modeled trajectories as a ``spatio-temporal language on a map,'' shifting the task from data distribution learning of traditional methods to instruction-based text generation \citep{Choi2020TrajGAILGU, feng2025agentmove}.
They leverage powerful contextual understanding and reasoning capabilities, offering advantages in semantic interpretability and versatility to different scenarios \citep{wang2024large}.

Current LLM-based methods explore various modeling strategies for generating realistic trajectories.
One line of work studies single-stage direct prompting.
For example, \citet{wang2023would,feng2024move} concatenate available information such as long and short-term check-ins and instruct an LLM to jointly model user preferences, geospatial distance, and sequential dynamics, yielding coherent trajectories.
Profile augmented modeling with a multi-stage pipeline is also a prominent research direction.
For instance, \citet{wang2024large,NEURIPS2024_3fb6c52a,Ju2025TrajLLMAM} first apply an LLM to infer semantic profiles from user histories such as personas and travel motivations, and then condition generation on these high-level abstractions to produce personalized trajectories.

\begin{figure}[!t]
\includegraphics[width=0.99\textwidth]{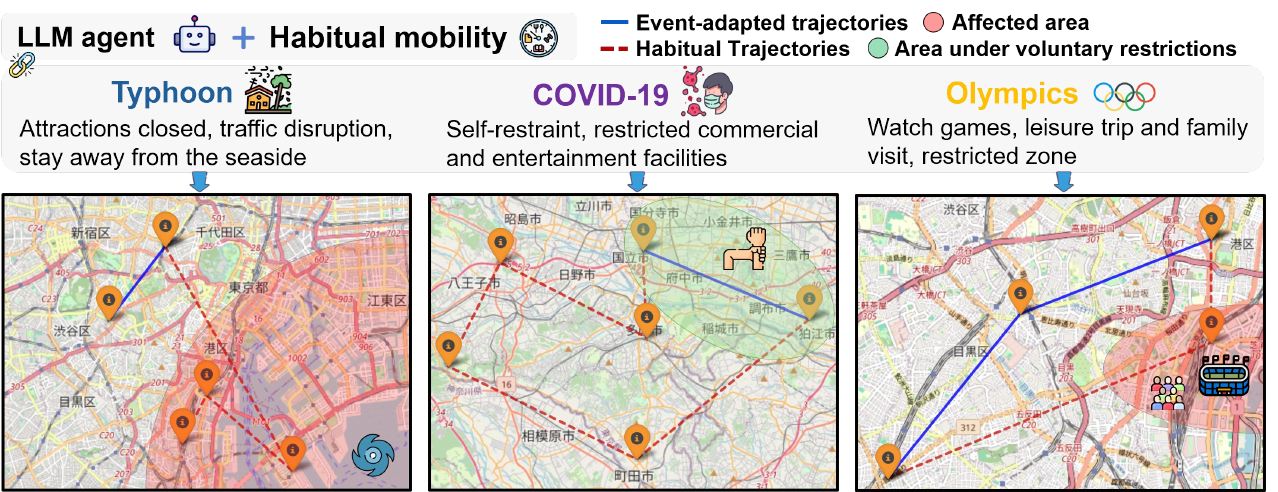}
\centering
\caption{
Event-driven mobility generation by LLMs, which incorporates event context to capture real-world human mobility on three different events: (1) Typhoon: evacuation from seaside, (2) COVID-19 Pandemic: self-restraint, and (3) Olympics: restricted zones and traffic jam.
}
\label{event-driven example}
\end{figure}

Although these methods achieve remarkable success, they still suffer from two key weaknesses.
The first is \textbf{data scarcity leading to evaluation bias}.
These methods are developed and evaluated primarily on datasets dominated by non-event days (stable period), resulting in questionable reliability when modeling non-routine deviations caused by large-scale societal events (e.g., natural disasters, public-health emergencies) \citep{10778764}.
As \autoref{event-driven example} (a) illustrates, during a typhoon, travel shifts away from coastal areas and unnecessary commutes are eliminated.
Without reliable data to evaluate their performance in modeling these dynamics, the reliability of these models for downstream applications, such as emergency response planning and transportation management under stress, is severely reduced \citep{Li2017ForecastingSS}.
Another limitation is \textbf{lack of a mechanism to reconcile competing decisions}.
In these events, real-world human mobility combines habitual regularities with shock-induced deviation \citep{song2014prediction}.
As \autoref{event-driven example} (b) and (c) show, while overall mobility patterns are altered, event-adapted trajectories preserve visits to essential anchor points (shared nodes for both lines) of a user's routine, such as workplaces.
Current methods struggle to navigate this duality, producing trajectories that either default to habitual patterns or are dominated by event constraints.
Thus, an explicit reconciliation mechanism is needed to generate plausible trajectories.

\vspace{-1pt}

To tackle these challenges,
we first develop an event-centric dataset to provide the necessary empirical foundation for studying non-routine mobility.
It covers trajectories from over a thousand users in the Tokyo metropolitan area cross three large-scale societal events (COVID-19 Pandemic, Typhoon Hagibis, Tokyo Olympics) with distinct mobility effects, in addition to a normal period for baseline comparison.
\wy{Second, we introduce ELLMob, a self-aligned LLM framework that incorporates cognitive theory to shift generic self-alignment from error correction to conflict reconciliation, explicitly arbitrating between these competing decisions.}
Our key insight draws from Fuzzy-Trace Theory (FTT) \citep{REYNA19951}, which posits that \textit{gist}, the essential meaning distilled from information, guides decisions under uncertainty.
Event-driven mobility naturally fits this perspective, as individuals weigh habitual patterns against event-imposed constraints.
Crucially, FTT reveals that gist can be linguistically expressed, enabling us to analyze the decision basis of LLMs. Building on these insights,
ELLMob extracts three forms of gist to capture competing decision rationales: \textit{pattern gist} (habitual tendencies) and \textit{event gist} (constraint requirements), along with the \textit{action gist} (LLM's current trajectory decision).
Through iterative alignment of these gists, ELLMob transforms trajectory decision into a traceable process where competitions are explicitly identified and reconciled, generating habitually grounded and event-responsive trajectories. 

Experiments show that on our event-centric dataset, existing methods produce trajectories that either default to routine patterns or overfit to event shocks (\autoref{exp_typ_covid}), leading to poor generation quality. In contrast, ELLMob effectively reconciles this duality, achieving the best performance across four metrics, and surpasses strongest baselines by an average of \textbf{46.9\%} across all three events (\autoref{tab:main_results}). Ablation studies confirm the critical role of gist-level reconciliation: incorporating cognitive-based self-alignment improves performance by an average of \textbf{69.5\%} over non-aligned variants, highlighting its necessity for event-driven mobility modeling. 
Our contributions are summarized as follows.
\begin{itemize}[leftmargin=*]
\item We construct the first event-centric human mobility dataset with detailed semantic information, providing a foundation for studying the non-routine deviations caused by societal events.
\item We provide the first empirical evidence that current LLM-based methods struggle to model human mobility under societal events, revealing a critical research gap.
\wy{ \item We propose an FTT-inspired framework that constructs decision variables to explicitly reconcile conflicts between habitual patterns and event constraints, enabling traceable decision-making.}
\item ELLMob achieves state-of-the-art (SOTA) performance across all evaluated scenarios, demonstrating its effectiveness in generating plausible human mobility behaviors.
\end{itemize} 

\section{Related Work}

\vspace{-6pt}
\subsection{Human Mobility Generation}

\vspace{-6pt}
The task of human mobility generation focuses on synthesizing realistic trajectories \citep{Sun2023SynthesizingRT,Gong2023TwoStageTG}. 
Early deep learning methods applied sequential models like LSTMs and attention-based RNNs to predict temporal dependencies and personal preferences \citep{10.1162/neco.1997.9.8.1735,Kulkarni2017GeneratingSM,Gao2017IdentifyingHM,Gao2018TrajectorybasedSC,Wang2018CompleteUM,10.1145/3178876.3186058,Luo2021STANSA}. 
To improve trajectory fidelity, subsequent research shifted to deep generative models, including VAEs~\citep{huang2019variational}, GANs~\citep{Choi2020TrajGAILGU,Wang2021LargeSG,Zhao2023GeneratingST,Jia2024DomainKnowledgeEG}, and diffusion models~\citep{zhu2023difftraj,NEURIPS2023_4786c0d1,Chu2024SimulatingHM}, which excel at generating high-resolution location sequences.
The emergence of LLMs introduced a new approach, re-framing trajectory generation as a sequence generation task conditioned on contextual prompts \citep{Xue2022LeveragingLF, wang2023would, feng2024move}.
However, the defined task for all these preceding models has been to simulate the routine activity trajectories of users. 
Their ability to generate faithful trajectories under sudden, non-stationary conditions such as disasters or public health crises therefore remains unknown, compromising their real-world application. Our work addresses this deficiency by defining the task of event-driven human mobility generation.

\vspace{-6pt}
\subsection{LLM for Human Mobility Modeling}
\vspace{-6pt}

LLMs are currently applied across a range of human mobility modeling strategies \citep{wang2023would, feng2024move, wang2024large, Tang2024InstructionTuningLE, Zhang2024MobGLMAL, beneduce2025large}. 
For example, \citet{wang2023would} incorporated both long- and short-term dependencies from historical mobility data into LLMs to generate the next visiting location.
\citet{LIANG2024102153} used an LLM with historical mobility Origin-Destination data to generate travel demand during public events at the Barclays Center.
\citet{wang2024large} integrated diverse user contexts, such as activity patterns, motivations, and profiles, into an LLM to generate more interpretable daily trajectories.
Existing LLM-based methods fail to reconcile competing objectives during events: they either blindly follow habitual patterns or event constraints, making them unable to effectively adapt to sharp mobility behavioral changes driven by events \citep{Luo2024DecipheringHM}.
In contrast, ELLMob is cognitive theory-driven and incorporates a self-aligned mechanism that iteratively adjusts generated trajectories, shifting the generation goal from maximizing statistical likelihood to cognitive plausibility.

\vspace{-4pt}
\section{Problem Definition}
\vspace{-6pt}

In this section, we define the terminology and formulate the event-driven trajectory generation problem.
A \textbf{trajectory} $\tau$ is a time-ordered sequence of visited places, represented as \begin{footnotesize}$\{(p_0, t_0), (p_1, t_1), \dots, (p_n, t_n)\}$\end{footnotesize}, where each tuple $(p_i, t_i)$ denotes a visit to place $p_i$ at time $t_i$.
\textbf{Event Context}, denoted $E_{ctx}$, is structured data describing the exogenous shock associated with a specific event $c$.
For a user $u$, we partition their historical trajectories within a pre-event window $W^{\mathrm{pre}}(c)$ into two disjoint sets based on a short-term duration $T_{\mathrm{short-term}}$ relative to the start time of event $t_c$: \textbf{long-term trajectories} \begin{footnotesize}$D_{\mathrm{long-term}}^{(u)} = \{\tau^{(u,d)} \mid d < t_c - T_{\mathrm{short-term}}\}$\end{footnotesize} and \textbf{short-term trajectories} \begin{footnotesize}$D_{\mathrm{short-term}}^{(u)} = \{\tau^{(u,d)} \mid d \ge t_c - T_{\mathrm{short-term}}\}$\end{footnotesize}.
These two datasets jointly characterize the prior mobility patterns of the user. 
The objective of this task is to develop a generative model that generates the event-driven trajectory: \begin{footnotesize}$F: (D_{\mathrm{long-term}}^{(u)}, D_{\mathrm{short-term}}^{(u)}, E_{ctx}) \mapsto \tau$\end{footnotesize}.

\vspace{-6pt}
\section{Event Human Mobility Data}
\vspace{-6pt}

To develop and evaluate the performance of models in capturing mobility shifts under varying events, we construct a dataset from Tokyo trajectories collected via Twitter and Foursquare (2019-2021).
It encompasses three distinct events selected to represent a spectrum of societal conditions, as well as a normal period to establish a baseline.
Detailed specifications are provided in \autoref{tab:event_scenarios}.
\begin{table}[h!]
\centering
\small 
\setlength{\tabcolsep}{3mm}{
    \caption{Specifications for experimental evaluation scenarios.}\vspace{-0.2cm}
    \label{tab:event_scenarios}
    \begin{tabular}{llll}
    \toprule
    \textbf{Event} & \textbf{Event Period} & \textbf{Pre-Event Period} & \textbf{Description} \\ \midrule
      Typhoon Hagibis &
      2019-10-12 $\sim$ 10-13 &
      2019-08-13 $\sim$ 10-11 &
      Natural disaster. \\
      COVID-19 Pandemic &
      2020-04-07 $\sim$ 04-13 &
      2020-02-07 $\sim$ 04-06 &
      Public health emergency. \\
      Tokyo 2021 Olympics &
      2021-07-23 $\sim$ 07-29 &
      2021-05-24 $\sim$ 07-22 &
      Pandemic-era large event. \\
      Normal Period &
      2019-09-01 $\sim$ 09-30 &
      2019-07-03 $\sim$ 08-31 &
      Regular urban mobility. \\ \bottomrule
\end{tabular}
}
\end{table}

For prolonged events (COVID-19 Pandemic and Tokyo Olympics), we focus on the first seven days to capture pronounced behavioral shifts. 
COVID-19 Pandemic began with the State of Emergency of Japan.
30-day window for normal period is used to establish a robust baseline of typical mobility, averaging out weekly fluctuations.
Pre-event period (two months) acts as training data for deep learning baselines and a source for user pattern extraction for LLM-based baselines.
Following collection process described in \autoref{data_collection}, we curated a dataset of 1,100 users who exhibited consistently dense check-in activity throughout the study period.
Each sample includes user ID, geographical coordinates, subcategory, subcategory ID, category, timestamp, and a comment, as shown in \autoref{mobilitydatasample}.
\autoref{tab:dataset_stats} shows key statistics of this dataset.
\autoref{tab:data_dimensions} compares data dimensions across major mobility datasets (GeoLife \citep{10.1145/1526709.1526816}, Gowalla \citep{10.1145/2020408.2020579}, Foursquare \citep{6844862}, and Yelp \citep{asghar2016yelp}), \wy{revealing that ours cover all standard mobility dimensions.
To the best of our knowledge, our dataset is the first to cover a broad spectrum of distinct event types (long-term vs. short-term, diverse semantics) with continuous, dense pre- and during-event trajectories, enabling the precise analysis of behavioral transitions during societal shifts.
}

\begin{table}[h!]
\centering

\vspace{-5pt}
\begin{minipage}{.5\textwidth}
    \centering
    \setlength{\tabcolsep}{0.40mm} 
    \captionsetup{width=.96\linewidth} 
    \caption{Dataset statistics by scenario, detailing the counts of check-ins, unique POIs, and POI categories, with per-scenario totals.}
    \label{tab:dataset_stats}
    \begin{tabular}{@{}lccc@{}} 
    \toprule
    \textbf{Event} & \textbf{Check-ins} & \textbf{POIs} & \textbf{Subcat.} \\
    \midrule
    Typhoon Hagibis & 4,330 & 3,284 & 334 \\
    COVID-19 Pandemic & 8,448 & 4,629 & 330 \\
    Tokyo 2021 Olympics & 16,069 & 9,586 & 467 \\
    Normal Period & 104,781 & 36,487 & 655 \\
    Agg. Pre-Event & 641,747 & 125,710 & 764 \\
    \bottomrule
    \end{tabular}
\end{minipage}
\begin{minipage}{.52\textwidth}
    \centering
    \setlength{\tabcolsep}{1.45mm}{
    \captionsetup{width=0.90\linewidth}
    \caption{T, L, C, TC, N, and E denote time, location, subcategory, comments, normal period and explicit event annotation, respectively.}
    \label{tab:data_dimensions}
    \begin{tabular}{lcccccc}
    \toprule
    \textbf{Dataset} & \textbf{T} & \textbf{L} (lat,lon) & \textbf{C} & \textbf{TC} & \textbf{N} & \textbf{E} \\
    \midrule
    GeoLife & \cmark & \cmark & \xmark & \xmark & \cmark & \xmark \\
    Gowalla  & \cmark & \cmark & \xmark & \xmark & \cmark & \xmark \\
    Foursquare & \cmark & \cmark & \cmark & \xmark & \cmark & \xmark \\
    Yelp & \cmark & \cmark & \cmark & \cmark & \cmark & \xmark \\
    \textbf{Ours} & \cmark & \cmark & \cmark & \cmark & \cmark & \cmark \\
    \bottomrule
    \end{tabular}
    }
\end{minipage}

\end{table}
\vspace{-5pt}

\begin{figure}[!h]
\includegraphics[width=1\textwidth]{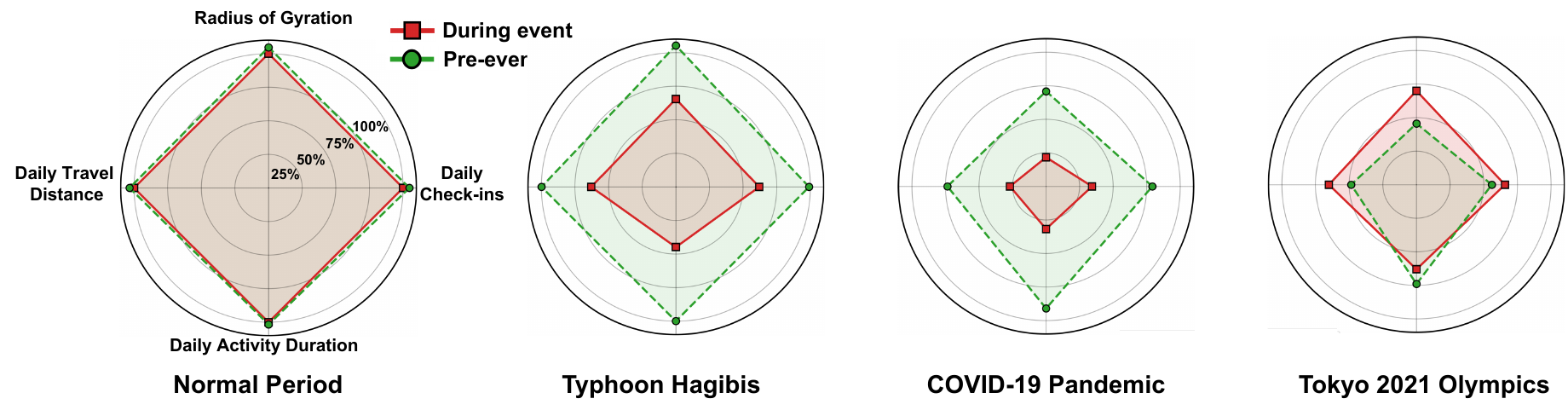}
\centering
\caption{Normalized radar charts of four mobility metrics on pre-event and during-event patterns.}
\label{matrix_statistics}\vspace{-0.3cm}
\end{figure}

To quantitatively ground this study, we present a statistical analysis to reveal distinct impacts of each event on collective mobility.
We adopt four widely-used metrics from the human mobility work \citep{pappalardo2015returners,alessandretti2020scales}: daily check-ins, capturing activity intensity, the radius of gyration and total travel distance, measuring spatial extent and volume, respectively, and the daily activity duration, quantifying the temporal span. 
Results are shown in \autoref{matrix_statistics}.
Specifically, COVID-19 Pandemic and Typhoon Hagibis significantly suppressed the scope and frequency of movement. 
In contrast, Olympics reversed suppressive trend in activity of COVID-19 Pandemic.
\vspace{-6pt}

\vspace{-6pt}
\section{Methodology}
\vspace{-6pt}

\subsection{Event Schema Construction}
A challenge in event-driven mobility generation is that real-world events are typically described in lengthy, free-form text (e.g., news reports and policy documents), which often leads the LLM to overlook critical information during trajectory generation \citep{NEURIPS2024_f169ec4d,NEURIPS2024_71c3451f}.
To address this, an event schema construction step is introduced to transform raw event narratives into a structured representation that explicitly outlines the event's impact on population mobility patterns.
The event schema is designed around four distinct but complementary aspects that collectively cover the key information required to assess mobility changes:

\begin{itemize}[
topsep=0pt,        
partopsep=0pt,     
itemsep=2pt,       
parsep=0pt,        
leftmargin=1em     
]
\item \textit{Event profile}: Records the fundamental elements of the event (e.g., type, name, occurrence time, and affected regions). This provides an anchor for spatio-temporal alignment.
\item \textit{Intensity and scale}: Quantifies key metrics of the event's severity, such as wind speed and amount of precipitation, to inform travel risk assessment.
\item \textit{Infrastructure and service impact}: Describes the operational status of critical resources (e.g., transportation, public venues), defining the physical constraints on mobility.
\item \textit{Official directives}: Captures governmental orders and recommendations (e.g., a request for residents to avoid non-essential travel), including their applicable populations and geographic scope, ensuring generated trajectories refer to policy mandates.
\end{itemize} 

An LLM is leveraged to process the raw event text $E_c$ into a structured key-value format, referred to as event context $E_{ctx}$, which subsequently serves as the input to the trajectory generation task.
The prompt is provided in \autoref{event_schema} and the generated contents are presented in \autoref{appendix:prompt_schema_sample}. 

\subsection{Self-aligned LLM Framework}
\begin{figure}[!t]
\includegraphics[width=0.9\textwidth]{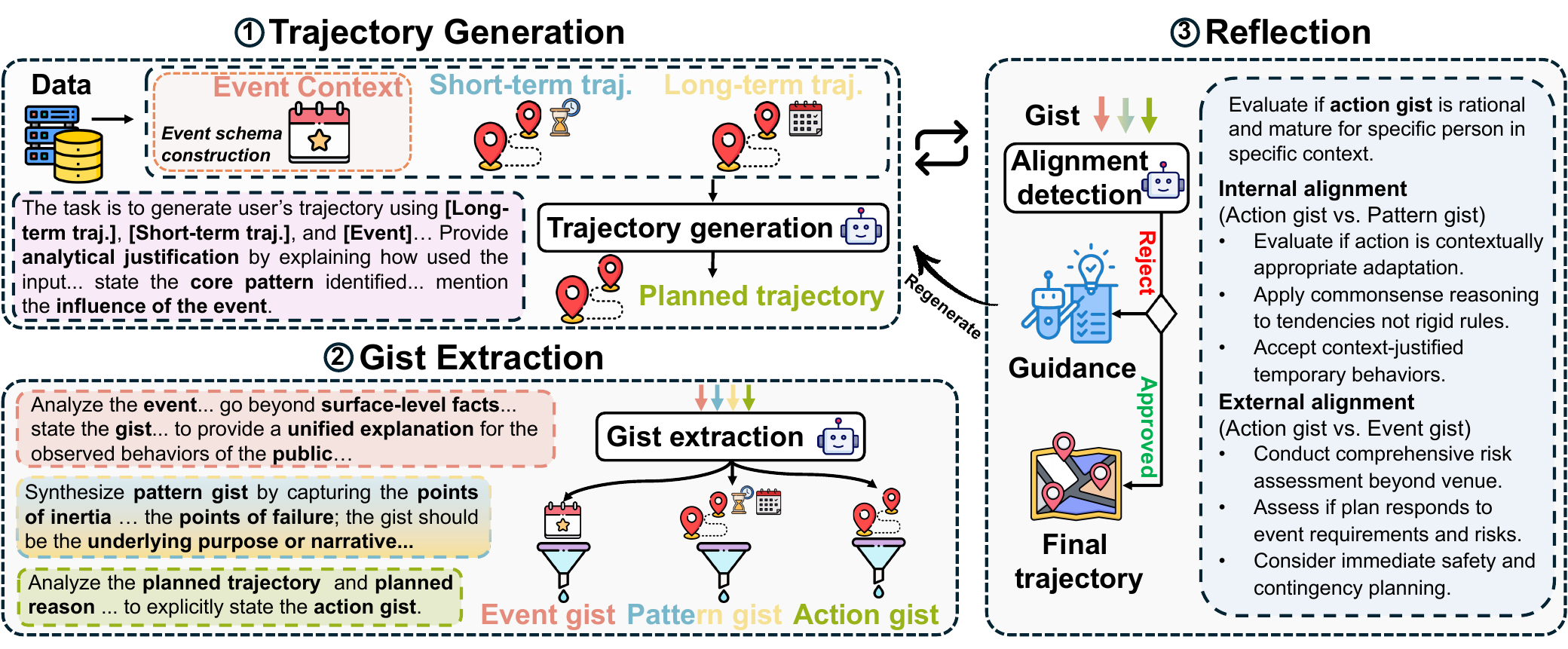}
\centering
\vspace{-0.2cm}
\caption{The ELLMob framework architecture comprising three interconnected modules: Trajectory Generation, Gist Extraction, and Reflection-based alignment.}
\label{model_structure}\vspace{-0.5cm}
\end{figure}

In \autoref{model_structure}, we propose ELLMob, a cognitive theory-driven framework that employs an iterative refinement process to reconcile competition between a user's habitual patterns and event constraints. 

\vspace{-0.2cm}
\subsubsection{Theory of Planned Mobility Behavior}
\vspace{-0.2cm}
Fuzzy-Trace Theory (FTT) \citep{REYNA19951} provides a cognitive perspective on decision-making under uncertainty, emphasizing that decisions are driven by \emph{gist}, which refers to the bottom-line (essential) meaning of information rather than \textit{verbatim} details.
A classic example is evacuation, where the action is driven not by the exact probability that a tsunami will strike (e.g., 15\%) but by the gist that the risk is ``high.''
According to FTT, gist can be linguistically expressed, making the decision-making basis transparent.
In event-driven mobility generation, uncertain disruptions such as natural disasters or epidemics require the model to navigate between two independent decision bases: adhering to habitual mobility routines or complying with event-imposed constraints.
Existing LLM-based methods lack an explicit mechanism for arbitrating between these competing decisions, leaving their rationale difficult to audit or control and tending to follow only one gist.
By extracting the gist underlying these decisions, we expose the model’s decision basis and resolve competition in a transparent manner.
We extract relevant gists: \textit{Pattern Gist} and \textit{Event Gist}, corresponding to two independent decision bases, and \textit{Action Gist}, representing LLM's tentative plan.
\begin{itemize}[
topsep=0pt,        
partopsep=0pt,     
itemsep=2pt,       
parsep=0pt,        
leftmargin=1em     
]
\item \textit{Pattern gist}: A representation of the essential tendencies distilled from the user’s habitual mobility patterns, reflecting stable movement routines. 
\item \textit{Event gist}: A representation of the essential tendencies distilled from contextual constraints, capturing constraints or incentives imposed by external events.
\item \textit{Action gist}: A representation of LLM’s immature mobility decision, extracted from the candidate trajectory during planning.
\end{itemize} 

We heuristically formalize these concepts as structured representations, where each gist is derived by assessing the relevant source data along a set of core attributes, which is illustrated in \autoref{tab:gist_definitions}.
Building on this, we propose a reflection module in which the LLM audits the alignment of these gist.
This alignment process explicitly identifies conflicts that are then resolved through guided refinement to ensure that the final generated trajectory is grounded in a unified decision basis.
\wy{Notably, FTT offers a architecture design basis. 
It motivates a multi-gist decision framework, guides mapping heterogeneous inputs into a unified gist space for consistent alignment, and drives the use of interpretable bottom-line attributes over arbitrary features. 
Ablation study is provided in \autoref{sec:ftt_justification}.}

\vspace{-0.2cm}
\begin{table}[h]
\centering
\footnotesize 
\caption{A set of defined core attributes that guide the gist extraction from source information.}\vspace{-0.25cm}
\label{tab:gist_definitions}
\resizebox{0.95\linewidth}{!}{
\begin{tabularx}{\linewidth}{@{}l|l>{\raggedright\arraybackslash}X >{\raggedright\arraybackslash}X @{}} 
\hline
\textbf{Gist Type} & \textbf{Attribute} & \textbf{Description} & \textbf{Example} \\ 
\hline
\multirow{5}{*}{Pattern Gist} 
& \multirow{1}{*}{Core Behavior} & \multirow{1}{*}{The dominant pattern of action.} & \multirow{1}{*}{Daily commute to a office.} \\ 
\cline{2-4}
& \multirow{2}{*}{Points of Inertia} & Deeply embedded, non-negotiable components. & Returning home to a specific neighborhood at night. \\
\cline{2-4}
& \multirow{2}{*}{Points of Fracture} & Critical dependencies and single points of failure. & Reliance on a single train line that might be suspended. \\
\hline
\multirow{7}{*}{Event Gist} 
& \multirow{2}{*}{Primary Intent} & Core implication of the event for mobility decisions. & High risk outdoors, strong incentive to stay home. \\
\cline{2-4}
& \multirow{2}{*}{Behavioral Implications} & Survival, social dynamics, and compliance. & Evacuation from coastal areas, seeking indoor shelter. \\
\cline{2-4}
& \multirow{2}{*}{Risk-Reward Calculus} & A cost-benefit analysis of the response to event risks. & Risk of injury outweighs reward of a non-essential outing. \\
\hline
\multirow{7}{*}{Action Gist} 
& \multirow{2}{*}{Primary Intent} & Main purpose driving this trajectory choice. & To get essential supplies from a nearby store. \\
\cline{2-4}
& \multirow{2}{*}{Habit Adherence} & Degree of preservation in habitual patterns. & Low; this trip deviates from the usual work commute. \\
\cline{2-4}
& \multirow{2}{*}{Event Compliance} & Trajectory's level of adherence to event constraints. & High; the trip is short and avoids dangerous areas. \\
\hline
\end{tabularx}
}
\end{table}

\vspace{-14pt}
\subsubsection{Reflection-based Alignment} 
\vspace{-0.1cm}
\wy{
We replace single-pass decoding with an iterative reflect-refine loop that externalizes the model's decision basis for transparent reasoning. Moreover, unlike generic self-alignment approaches that primarily correct errors such as hallucinations, our mechanism targets the decision-making dilemma inherent in event-driven mobility scenarios.} Our alignment performs in two stages:

\textbf{Alignment Auditing.} 
This process is dedicated to rigorously auditing the plausibility of a planned trajectory.
Each candidate trajectory is checked along two binary dimensions:
\textit{Internal alignment} is to ascertain whether the planned trajectory reflects a coherent expression of the user's intrinsic habitual mobility patterns and current behavioral tendencies.
\textit{External alignment} determines if the planned trajectory represents a rational and compliant response to the constraints and implications of the event.
A trajectory is accepted only if both criteria are satisfied. 
The auditor outputs two binary judgments, accompanied by concise rationales that indicate any violated criterion and its cause.

\textbf{Corrective Refinement.} Should a planned trajectory fail to satisfy the criteria of either the internal or external alignment audit, ELLMob initiates a corrective refinement loop. 
During this loop, the precise reasons for the audit failure are provided as feedback to the trajectory generator, guiding it to regenerate a revised trajectory that explicitly addresses the identified semantic misalignments and logical flaws. 
This loop repeats up to a maximum of $K$ iterations.
A trajectory that satisfies both criteria within the $K$-step budget is accepted as a final trajectory.
\wy{In the rare event that constraints remain unmet after $K$ iterations, the system executes a fallback strategy. 
It accepts the last validated trajectory available in the agent’s buffer and reports unmet constraints to ensure transparency.
}

To clearly understand the ELLMob, we show the overall procedure in pseudo-code form \autoref{algorithm} with complete contents of all related prompts provided in \autoref{completeprompt}.

\section{Experiments}

\vspace{-0.2cm}
\subsection{Experimental Setup}
\label{exp_details}

\textbf{Baselines.}
ELLMob is evaluated against two types of baselines:
\textit{1) deep learning-based methods} which include predictive models: 
LSTM \citep{10.1162/neco.1997.9.8.1735}, DeepMove \citep{10.1145/3178876.3186058}, GETNext \citep{10.1145/3477495.3531983}, and MHSA \citep{HONG2023104315};
and generative models: TrajGAIL \citep{Choi2020TrajGAILGU} and DiffTraj \citep{zhu2023difftraj}.
\textit{2) LLM-based models}: LLM-MOB \citep{wang2023would}, LLM-Move \citep{feng2024move}, LLMOB \citep{wang2024large}, LLM-ZS \citep{beneduce2025large}.
\wy{For fairness, the input event information remains consistent across all LLM-based methods. Specifically, detailed event descriptions (including type, time, location, and constraints) are integrated as natural language context at the beginning of each prompt, ensuring that all baselines have equal access to the event information despite differences in their specific prompt designs.}

\textbf{Evaluation Metrics.}
\textit{Step Interval (SI)}.
The time between consecutive activities, defined as
$\text{SI}_t = \tau_{t+1} - \tau_t$, 
where $\tau_t$ denotes the timestamp at step $t$;
\textit{Step Distance (SD)}.
The distance between consecutive locations, defined as
$\text{SD}_t = \| l_{t+1} - l_t \|_2$, 
where $l_t \in \mathbb{R}^2$ denotes the location at step $t$.
\textit{Category Distribution (CD)}. This metric captures the distribution of activity types. 
To calculate it, we aggregate the total number of visits $N(c_k)$ for each location category $c_k$.
\textit{Spatial Grid Distribution (SGD)}.
It captures the population-level spatial footprint of activities.  
All visited locations are discretized onto a fixed $S \times S$ grid covering the Tokyo metropolitan area, with visit counts accumulated per grid cell.  
To mitigate sparsity, following \citet{ouyang2018non,feng2020learning}, the top $25\%$ frequently visited cells are retained for evaluation.
For each of the four metrics, we form a distribution from the generated trajectories and compare it against the ground truth distribution using the Jensen-Shannon Divergence (JSD), following \citet{zhu2023difftraj,wang2024large}.


\wy{
\textbf{Implementation Details.} 
We primarily use GPT-4o-mini (2025-01-01-preview) \citep{achiam2023gpt} as the backbone for its capability–cost balance, with additional LLM evaluations reported in \autoref{cross_llms}.
Following \citet{wang2024large}, we set the temperature to 0.1 to curb randomness, Top-p to 1, and model trajectories at a 10-minute resolution.
Grid size parameter $S$ is set to 10. 
$K$ is set to 3 to balance refinement quality and inference cost based on the parameter study in \autoref{parameter_sensitive_iteration}.
A stability analysis verifying result consistency is provided in \autoref{sec:stability}.
}

\begin{table}[t]
\caption{Comparison of different methods under three events. 
Performance is evaluated by JSD across four dimensions with the best performance highlighted in \textbf{bold}.}
\vspace{-0.2cm}
\label{tab:main_results}
\centering
\small
\setlength{\tabcolsep}{0.62mm}{
    \begin{tabular}{l|cccc|cccc|cccc}
    \toprule
    \multirow{2}{*}{\textbf{Models}} & \multicolumn{4}{c|}{\textbf{Typhoon Hagibis}} & \multicolumn{4}{c|}{\textbf{COVID-19 Pandemic}} & \multicolumn{4}{c}{\textbf{Tokyo 2021 Olympics}} \\
     & SI↓ & SD↓ & CD↓ & SGD↓ & SI↓ & SD↓ & CD↓ & SGD↓ & SI↓ & SD↓ & CD↓ & SGD↓ \\
    \midrule
    LSTM & 0.1336 & 0.1039 & 0.0555 & 0.1111 & 0.1928 & 0.1047 & 0.1300 & 0.2571 & 0.1147 & 0.0651 &	0.0598 & 0.0634 \\
    DeepMove & 0.1697 & 0.0826 & 0.0266 & 0.0759 & 0.1838 & 0.0834 & 0.0423 & 0.1688 &0.1667 &	0.0492 & 0.0587 & 0.0555\\
    GETNext & 0.3031 & 0.2007 & 0.0274 & 0.1037 & 0.2891 & 0.2241 & 0.0142 & 0.1354 & 0.2701  &	0.1473 & 0.0176 & 0.1204\\
    MHSA  & 0.1430 & 0.1815 & 0.0118 & 0.0711 & 0.2180 & 0.3083 & 0.0254 & 0.0437 & 0.1815 & 0.2013 & 0.0120 & 0.0525 \\
    TrajGAIL & 0.1034 & 0.3591 & 0.0155 & 0.0275 & 0.1600 & 0.3557 & 0.0195 & 0.0444 & 0.0863 &	0.2913 & 0.0121 & 0.0104\\
    DiffTraj & 0.1271  & 0.2450 & 0.0385 & 0.0761 & 0.1405 & 0.2766 & 0.0554 & 0.0454 & 0.0732 & 0.2171 & 0.0342 & 0.0282 \\
    LLMOB & 0.0949 & 0.1195 & 0.0123 & 0.0256 & 0.1013 & 0.1051 & 0.0186 & 0.0286 & 0.0973	& 0.0274 &	0.0110 & 0.0051 \\
    LLM-MOB & 0.1214 & 0.0468 & 0.0285 & 0.0344 & 0.1166 & 0.0532 & 0.0234 & 0.0353 & 0.1047 &	0.0286 & 0.0085 & 0.0052 \\
    LLM-Move & 0.1267 & 0.0392 & 0.0136 & 0.0303 & 0.1408 & 0.0567 & 0.0127 & 0.0503 &  0.1967 &	0.0298 & 0.0101 & 0.0057 \\
    LLM-ZS & 0.1574 & 0.1348 & 0.0153 & 0.0724 & 0.1146 & 0.0576 & 0.0552 & 0.0570 & 0.0938 & 0.0330 & 0.0132 & 0.0052 \\
    \midrule
    ELLMob & \textbf{0.0642} & \textbf{0.0200} & \textbf{0.0041} & \textbf{0.0173} & \textbf{0.1003} & \textbf{0.0444} & \textbf{0.0080} & \textbf{0.0268} & \textbf{0.0617} &	\textbf{0.0061} &\textbf{0.0022} &	\textbf{0.0035}\\
    w/o I.A.\&E.A.  & 0.1304 & 0.1270 & 0.0139 & 0.0723 & 0.2331 & 0.1077 & 0.1190 & 0.0733 & 0.1465 &	0.0340 &	0.0093 &	0.0095 \\
    w/o I.A. & 0.0835 & 0.0720 & 0.0135 & 0.0436 & 0.1235 & 0.0950 & 0.1053 & 0.0300 & 0.1355	& 0.0316 &	0.0088 &	0.0086 \\
    w/o E.A. & 0.0680 & 0.0258 & 0.0077 & 0.0229 & 0.2237 & 0.0860 & 0.0283 & 0.0430 & 0.1392 &	0.0291 &	0.0083 &	0.0064 \\
    w/o Eve. Ext. & 0.0736 & 0.0273 & 0.0045 & 0.0227 & 0.2037 & 0.0741 & 0.0269 & 0.0405 & 0.0686 & 0.0213 & 0.0030 & 0.0041\\
    \bottomrule
    \end{tabular}
}\vspace{-14pt}
\end{table}

\vspace{-10pt}
\subsection{Quantitative Results}
\vspace{-6pt}

\autoref{tab:main_results} summarizes our main results, demonstrating ELLMob's consistent superiority across all event-driven settings. 
For instance, it improves the SI score by 32.3\% for Typhoon Hagibis and the SD score by 16.5\% for the COVID-19 Pandemic compared to the strongest baselines. 
We observe that LLM-based approaches generally outperform traditional deep learning models, particularly on spatial coherence metrics (SD, SGD), benefiting from their ability to integrate event context.
\wy{
Furthermore, to verify spatial generalizability beyond the Tokyo area, we extended evaluations to Osaka during the COVID-19 pandemic, with detailed results provided in \autoref{sec:regional_eval}.}

The ablation study further dissects our ELLMob's performance. Removing either the reflection module (w/o I.A.\&E.A.) or the event schema (w/o Eve. Ext.) consistently degrades performance. 
The two components of the reflection module show distinct roles: Internal alignment (w/o E.A.) provides foundational plausibility and external alignment (w/o I.A.) acts as a scenario-specific corrective. The importance of external alignment is particularly evident during the COVID-19 Pandemic, where its removal causes a catastrophic 132.4\% performance degradation in the SI score. 
This highlights its critical role in aligning with rational behaviors that substantially deviate from habitual patterns.

\vspace{-5pt}
\subsection{Model Analysis}
\vspace{-3pt}

\begin{figure}[t]
\includegraphics[width=0.9\textwidth]{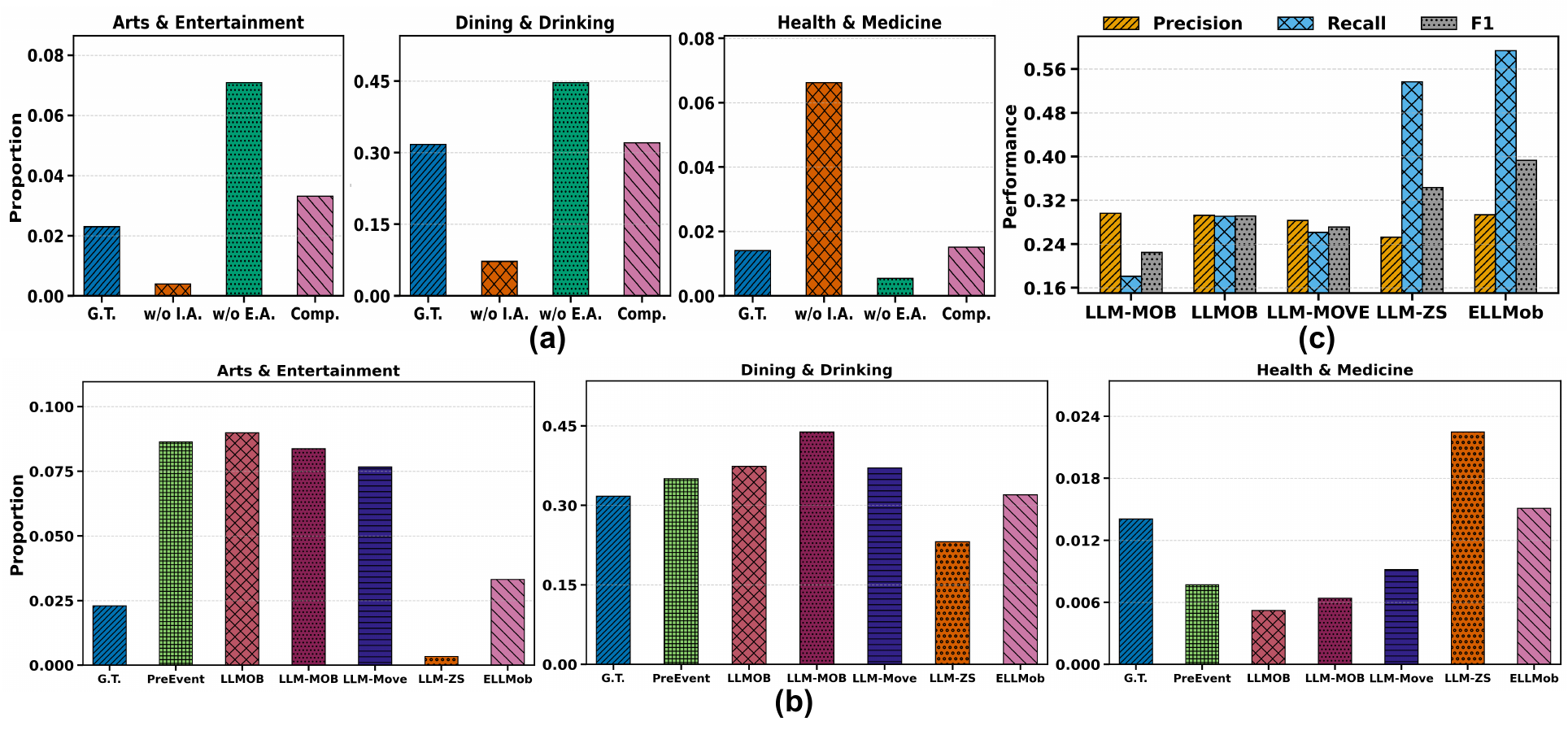}
\centering
\vspace{-0.5cm}
\caption{(a) Comparison of generated activity proportions (relative to the total number of activities) during the COVID-19 Pandemic. Each chart contrasts the ground truth (G.T.), distribution with: Without internal alignment (w/o I.A.), without external alignment (w/o E.A.), and the complete model (Comp.).
(b) Comparison of three key activity categories distributions generated by ELLMob with various LLM-based baselines.
(c) Performance comparison on the active user prediction task.}
\label{exp_typ_covid}
\vspace{-0.5cm}
\end{figure}

\textbf{Analysis of Self-alignment.}
To dissect the distinct roles of internal and external alignment, we analyze the generated distributions of three sensitive top-categories (Arts \& Entertainment, Dining \& Drinking, and Health \& Medicine) within the COVID-19 Pandemic.
As shown in \autoref{exp_typ_covid} (a), removing either alignment leads to distinct failures. 
Lacking internal alignment, the model over-corrects for the event, generating an unrealistic surge in Health \& Medicine while excessively suppressing Arts \& Entertainment and Dining \& Drinking activities. 
Conversely, without external alignment, the model rigidly adheres to habitual patterns, producing the opposite failure.
Furthermore, \autoref{exp_typ_covid} (b) reveals that most LLM-based baselines default to habitual patterns (PreEvent) such as overestimating entertainment/dining while ignoring health-related travel, while LLM-ZS overcorrects by suppressing social activities entirely.
Both extremes demonstrate that these baselines are unable to reconcile habitual patterns with event constraints in trajectory decision.
The complete ELLMob successfully considers these two forces to produce a distribution that closely matches the ground truth, demonstrating the effectiveness of the self-alignment mechanism.

\textbf{Fundamental Decisions in Disasters.}
Accurately identifying individuals who travel during extreme weather is critical for targeted early warnings and effective emergency response.
We frame this as a binary classification task to identify a potentially high-risk cohort, which we define as the positive class of ``active'' users (at least one trip) during the typhoon.
We evaluate LLM-based baselines for their strong ability to incorporate event context.
As shown in \autoref{exp_typ_covid} (c), ELLMob achieves the highest F1-Score, driven by its superior recall of 59.3\% in identifying this ``active" high-risk population.
This effectiveness is likely attributed to the iterative alignment process, which enhances LLM's joint understanding of individual user mobility patterns and event constraints.

\begin{figure}[h]
\centering
\begin{minipage}{0.48\textwidth}
    \centering
    \includegraphics[width=\textwidth]{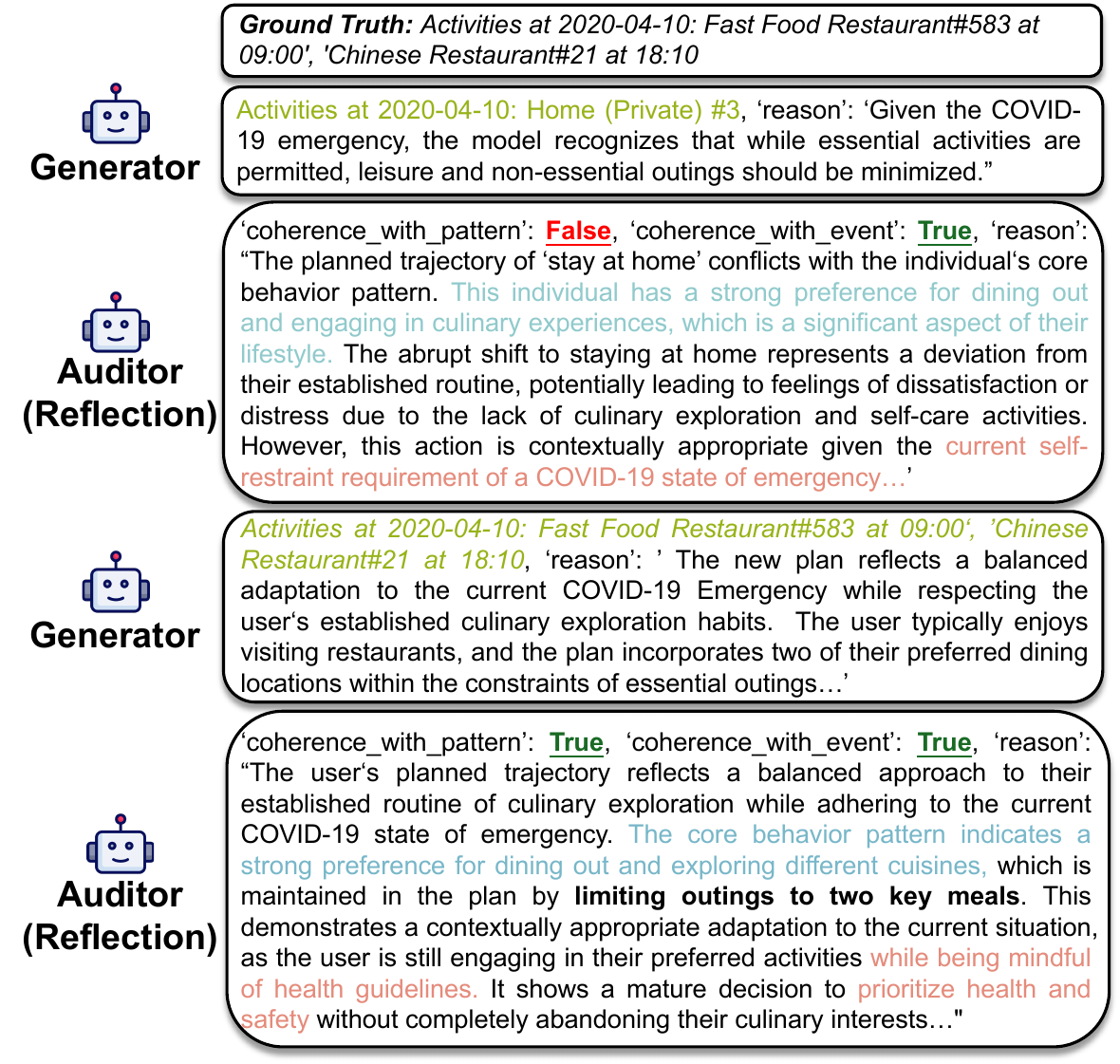}
    \captionof{figure}{A case study of ELLMob's workflow on the mobility of User No.003.}
    \label{case_study}
\end{minipage}
\hfill
\begin{minipage}{0.48\textwidth}
    \centering
    \captionof{table}{Comparison of different methods under the Normal period with the best in \textbf{bold}.}
    \label{table:normalday}
    \small
    \setlength{\tabcolsep}{1.1mm}{ 
        \begin{tabular}{l|cccc}
        \toprule
        \multirow{2}{*}{\textbf{Models}} & \multicolumn{4}{c}{\textbf{Normal period}} \\
         & SI↓ & SD↓ & CD↓ & SGD↓ \\
        \midrule
        LSTM & 0.1140 & 0.0696 & 0.0746 & 0.1499 \\
        DeepMove & 0.1423 & 0.0428 & 0.0300 & 0.0742 \\
        GETNext &  0.3071 & 0.1628 & 0.0126 & 0.0502\\
        MHSA & 0.1546 & 0.2346 & 0.0069 & 0.0269 \\
        TrajGAIL & 0.0953 & 0.3432 & 0.0035 & 0.0104 \\
        DiffTraj & 0.0748 & 0.1832 & 0.0361 & 0.0393 \\
        LLMOB & 0.1460 & 0.1007 & 0.0051 & 0.0045 \\
        LLM-MOB & 0.0654 & 0.0186 & 0.0059 & 0.0030 \\
        LLM-Move & 0.1836 & 0.0261 & 0.0067 & 0.0036 \\
        LLM-ZS & 0.0746 & 0.0311 & 0.0164 & 0.0027 \\
        \midrule
        ELLMob & \textbf{0.0496} & \textbf{0.0164} & \textbf{0.0025} & \textbf{0.0025} \\
        w/o I.A.\&E.A. &  0.0639 & 0.0210 & 0.0041 & 0.0032 \\
        w/o I.A. & 0.0545 & 0.0198 & 0.0026 & 0.0028 \\
        w/o E.A. & 0.0556 & 0.0201 & 0.0037 & 0.0028 \\
        \bottomrule
        \end{tabular}
    }
\end{minipage}
\end{figure}

\textbf{Case Study.}
To illustrate ELLMob’s reasoning process, \autoref{case_study} presents a case study of a user with a strong culinary exploration pattern during the COVID-19 Pandemic.   
The initially planned trajectory (stay at home) over-aligns with the event's public health constraints and is flagged by our reflection module for conflicting with the user's habitual patterns.
Guided by this internal feedback, the model iteratively refines the plan into a plausible trajectory that weighs both factors, ultimately limiting rather than eliminating dining outings. 
This case highlights the ELLMob’s ability to reconcile user patterns with event constraints to generate realistic behaviors that match the ground truth.

\vspace{-3pt}
\subsection{Evaluation on Routine Mobility}
\vspace{-3pt}
To assess ELLMob's generality, we evaluate its performance in routine scenarios. 
Since ELLMob is inherently event-driven, this evaluation requires defining a context for the normal period. 
We therefore classify target days as either weekdays or weekends, defining their respective contexts by the typical societal operating status (e.g., differences in business hours, public transport schedules). 
As shown in \textcolor[HTML]{fc6160}{Table} \ref{table:normalday}, ELLMob outperforms all baselines. This confirms that the ELLMob's alignment mechanisms are not narrowly tailored to disruptive events, but instead constitute a robust foundation that also excels in standard scenarios.

\wy{
\vspace{-3pt}
\subsection{Comparative Ablation on Alignment Strategies}
\vspace{-3pt}
Iterative-reflection methods such as Reflexion~\citep{Shinn2023ReflexionLA}, SELF-REFINE~\citep{Madaan2023SelfRefineIR}, and Air~\citep{liu-etal-2025-air} enhance LLM reasoning via self-correction, mainly targeting hallucinations or logical flaws in unstructured text.
However, event-driven mobility requires resolving conflicts between habitual inertia and event-induced constraints.
ELLMob introduces a methodological shift by grounding alignment in cognitive theory.
At representation level, it replaces unstructured trajectories with structured decision variables to disentangle drivers.
At decision level, rather than prompting the model to ``improve'' an answer, ELLMob employs dual-axis alignment to arbitrate competing objectives.
This ensures trajectory adjustments are cognitively grounded rather than surface-level, locally reasonable fixes.
To isolate the effectiveness of this design, we replaced ELLMob’s alignment module with each baseline strategy, while keeping all other settings identical.
As the results shown in \autoref{tab:reflection_results}, ELLMob outperforms these variants in all metrics. 
This validates the necessity of the proposed alignment strategy for event-driven human mobility generation. 
}

\vspace{-0.15cm}
\wy{
\begin{table}[h]
\caption{\wy{Comparison with iterative-reflection baselines with the best performance in \textbf{bold}.}}
\vspace{-0.2cm}
\label{tab:reflection_results}
\centering
{
\small
\setlength{\tabcolsep}{0.62mm}{
    \begin{tabular}{l|cccc|cccc|cccc}
    \toprule
    \multirow{2}{*}{\textbf{Models}} & \multicolumn{4}{c|}{\textbf{Typhoon Hagibis}} & \multicolumn{4}{c|}{\textbf{COVID-19 Pandemic}} & \multicolumn{4}{c}{\textbf{Tokyo 2021 Olympics}} \\
     & SI↓ & SD↓ & CD↓ & SGD↓ & SI↓ & SD↓ & CD↓ & SGD↓ & SI↓ & SD↓ & CD↓ & SGD↓ \\
    \midrule
    Reflexion   & 0.1106 & 0.1282 & 0.0841 & 0.0855 & 0.1685 & 0.0588 & 0.0146 & 0.0269 & 0.1741 & 0.0308 & 0.0378 & 0.0704 \\
    SELF-REFINE & 0.1979 & 0.0637 & 0.0135 & 0.0193 & 0.2122 & 0.1053 & 0.0344 & 0.0294 & 0.0826 & 0.0320 & 0.0073 & 0.0058 \\
    Air         & 0.0764 & 0.0710 & 0.0198 & 0.0204 & 0.1858 & 0.0454 & 0.0256 & 0.0291 & 0.0774 & 0.0441 & 0.0053 & 0.0035 \\
    ELLMob      & \textbf{0.0642} & \textbf{0.0200} & \textbf{0.0041} & \textbf{0.0173} & \textbf{0.1003} & \textbf{0.0444} & \textbf{0.0080} & \textbf{0.0268} & \textbf{0.0617} & \textbf{0.0061} & \textbf{0.0022} & \textbf{0.0035} \\
    \bottomrule
    \end{tabular}
}}
\end{table}
}
\vspace{-0.15cm}

\wy{
\begin{table}[h]
\centering

\caption{\wy{Computational Efficiency Analysis. The reported total token count includes both input and output tokens, and the overall cost is computed by accounting for their respective pricing rates.}}
\vspace{-0.2cm}
\label{tab:Comp_eff}
{
\begin{tabular}{lccc}
\toprule
\textbf{Model} & \textbf{Token Count} & \textbf{Inference Time (s)} & \textbf{Cost (USD)} \\
\midrule
LLMOB        & 1,271  & 10.12 & 0.00030 \\
LLM-MOB      & 3,954  & 3.72  & 0.00064 \\
LLM-MOVE     & 4,954  & 4.05  & 0.00078 \\
LLM-ZS       & 5,184  & 3.34  & 0.00080 \\
\midrule
Reflexion & 26,057 & 27.12 & 0.00417 \\
SELF-REFINE & 15,382 & 21.16 & 0.00258 \\
Air & 15,514 & 20.50 & 0.00260 \\
ELLMob       & 9,569  & 18.68 & 0.00170 \\
\bottomrule
\end{tabular}
}
\vspace{-3pt}
\end{table}
}

\wy{\subsection{Computational Efficiency}
We evaluated computational overhead via token consumption and inference latency on a \textit{per person per day} basis, with results in \autoref{tab:Comp_eff}.
Under GPT-4o-mini pricing, ELLMob uses 9,569 tokens and 18.68 seconds to generate one-day mobility for a single person, at \$0.00170.
While this multi-stage architecture entails additional overhead compared to single-pass models, it demonstrates superior efficiency relative to generic reflection baselines like Reflexion \citep{Shinn2023ReflexionLA}.
This efficiency might stem from the integration of FTT, where structured alignment provides targeted guidance to accelerate convergence and avoids the excessive resource cost of open-ended iterative refinement.
Independent user-level generation allows parallelization, ensuring city-scale simulations feasible.
}

\subsection{Visualization of Spatial Mobility Patterns}
\begin{figure}[ht]
\includegraphics[width=0.9\textwidth]{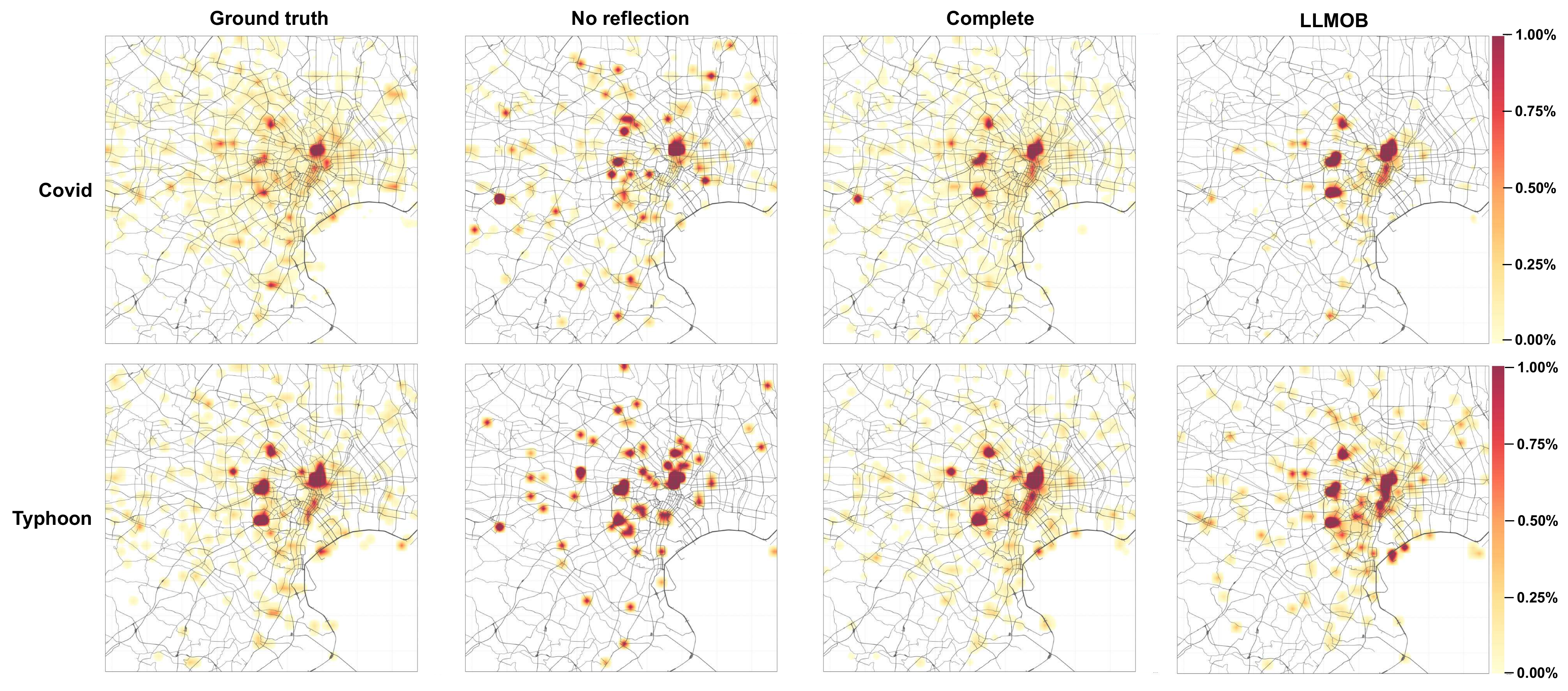}
\centering
\vspace{-0.3cm}
\caption{Spatial mobility patterns. Darker red in the heatmaps indicates higher visit frequency.}
\label{heatmap}
\end{figure}

\autoref{heatmap} presents heatmaps of the spatial mobility distribution during two high-impact events (Typhoon Hagibis and the COVID-19 Pandemic), comparing ground truth against ELLMob, its ablated version without the reflection module, and the strongest baseline LLMOB.
Both our ablated model  and the strongest baseline LLMOB exhibit the core limitations of single-pass generation with implicit trajectory decision, producing flawed spatial patterns: Excessive contraction during the typhoon and incomplete decentralization during the COVID-19 Pandemic. 
In contrast, ELLMob's reflection module uses iterative alignment to achieve a fine-grained understanding of both user patterns and event constraints, enabling it to reproduce realistic mobility patterns.

\vspace{-12pt}
\section{Conclusion}

This work addresses the critical challenge of modeling human mobility during large-scale societal events. 
We contribute a comprehensive event-centric dataset covering three major events in Tokyo and introduce ELLMob, a framework that explicitly reconciles competing mobility decisions through gist-based alignment. 
Through extensive experiments, ELLMob demonstrates substantial improvements over existing methods, enabling more reliable mobility generation for emergency planning and urban management applications.
\wy{However, we acknowledge that data from these platforms may introduce demographic biases, such as skewing towards younger users, which is a common limitation in LBSN research.
Future work will aim to incorporate more diverse data sources.}



\section*{Acknowledgments}
This work was supported by 
JSPS KAKENHI Grant JP24K02996, JP23K17456, JP23K25157, JP23K28096, JP25H01117 and JST CREST Grant JPMJCR21M2, JPMJCR22M2.

\bibliography{iclr2026_conference}
\bibliographystyle{iclr2026_conference}

\appendix
\renewcommand{\thetable}{A\arabic{table}}
\renewcommand{\thefigure}{A\arabic{figure}}
\setcounter{table}{1} 
\setcounter{figure}{1}

\section{Data Collection, Cleaning and Anonymization}
\label{data_collection}
\textbf{Data Collection.} The raw data used in this study are derived from Foursquare check-in records that users publicly synced to Twitter, which are accessible through the Twitter API. We first used the Twitter API to identify users who were active within a 100 km radius of Tokyo Station during April 2021, and then retrieved all of their tweets from 2019 to 2021. We subsequently identified Foursquare check-in tweets (auto-posted via the Foursquare→Twitter integration) and extracted the associated metadata, including the Point-of-Interest (POI) name, category, and subcategory\footnote{\url{https://docs.foursquare.com/data-products/docs/categories}}, geographic coordinates (latitude \& longitude), timestamp, and user-provided comment text. The extracted check-ins served as the raw dataset for this study. Importantly, these user-level records span 2019 to 2021, providing a longitudinal dataset that captures diverse social and environmental contexts, including but not limited to the Typhoon Hagibis, COVID-19 pandemic, and the Tokyo 2021 Olympics, which collectively form the foundation of this study.

\begin{figure}[!h]
\includegraphics[width=0.96\textwidth]{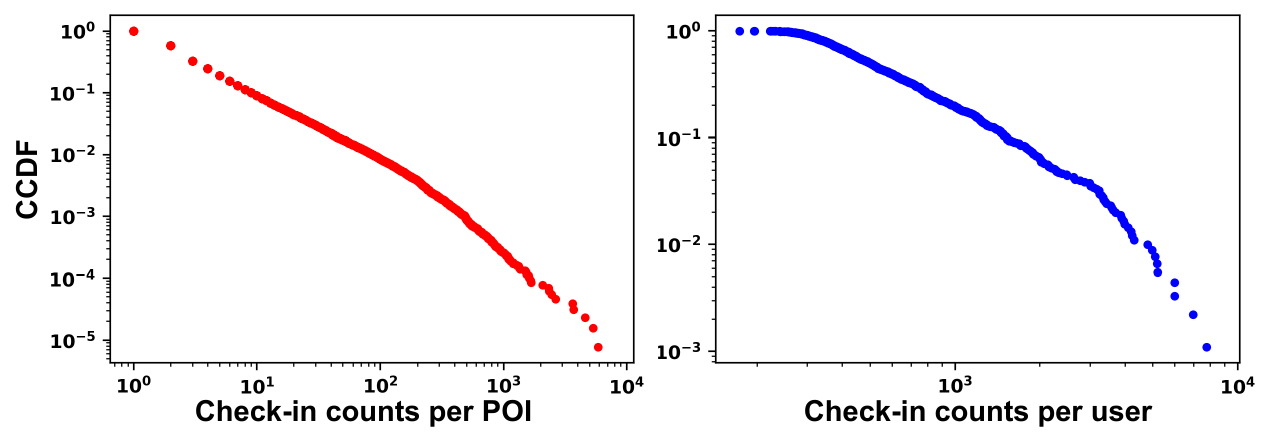}
\centering
\caption{Log-log plots of the Complementary Cumulative Distribution Function (CCDF) of check-in counts. 
Left part of the figure shows the distribution of check-ins per POI. 
Right part of the figure shows the distribution of check-ins per user. 
Both distributions exhibit a linear trend, characteristic of a power-law.}
\label{distribution_statistics}
\end{figure}

\noindent \textbf{Data Cleaning.} 
After obtaining the raw dataset, we performed several cleaning steps to improve data quality. First, we discarded users with missing check-ins for an entire year. Next, we parsed geographic coordinates to obtain prefecture information and assigned each user to their most frequently visited one. 
For example, users whose check-ins were primarily in Tokyo were labeled as ``Tokyo users''. 
We filtered the dataset to include only Tokyo users for two primary reasons: To ensure a homogeneous dataset by mitigating confounding variables from adjacent prefectures and to leverage the wide spectrum of mobility behaviors characteristic of a global mega city.
We retained only users with consistently dense check-in activity throughout the study period, yielding a final dataset of 1,100 users. 
As illustrated in \autoref{distribution_statistics}, the check-in distribution of these sampled users follows a power-law characteristic, which is consistent with real-world human activity patterns \citep{10.1145/3308558.3313635}. 

\noindent \textbf{Data Anonymization.} To protect privacy, we applied deterministic, one-way pseudonymization to all identifiers. Twitter user IDs were irreversibly mapped to integer surrogates, and POI identifiers were processed in the same manner. POI names (e.g., ``Yoshinoya Shinjuku'') were removed while retaining only category information (e.g., major category ``Dining and Drinking'' and subcategory ``Donburi Restaurant''). 
Precise time information was obfuscated to a resolution of one minute and all comments were translated into English and paraphrased.


\section{Mobility Data Sample}
\label{mobilitydatasample}
To provide a concrete illustration of the user activity trajectories, \autoref{tab:data_samples} presents an exemplary sequence from a single user. 
This sample highlights the data structure, integrating precise spatio-temporal features (latitude, longitude, time) with functional semantics (Location Name, Category), which forms the foundation for our mobility analysis.

\begin{table}[h!]
\centering
\scriptsize
\renewcommand{\arraystretch}{1.2}
\caption{A user's mobility sample, showcasing activity trajectories with spatio-temporal features and comments.}
\label{tab:data_samples}
\begin{tabularx}{\linewidth}{@{} l cc l l l l X @{}}
\toprule
\textbf{User} & \textbf{Lat.} & \textbf{Long.} & \textbf{POI ID} & \textbf{Subcategory} & \textbf{Category} & \textbf{Timestamp} & \textbf{Comments (Translated and rewritten)} \\
\midrule
0118 & 35.652 & 139.543 & 1003 & Home Appliance Store & Retail & 2020-04-07 18:33 & Chofu-chan: Oh no, an emergency declaration has been announced!! \\ 
\addlinespace[0.5ex]
0118 & 35.633 & 139.577 & 14932 & Clothing Store & Retail & 2020-04-10 19:16 & In response to coronavirus precautions, credit card transactions have transitioned to self-scanning! (This is what we’ve been aiming for.). \\
\addlinespace[0.5ex]
0118 & 35.632 & 139.577 & 4859 & Rail Station & Travel \& Transport & 2020-04-10 19:21 & Wait a minute? Isn't this considered close contact? \\
\bottomrule
\end{tabularx}
\end{table}

\section{Behavior distribution}

\begin{figure}[!h]
\includegraphics[width=0.96\textwidth]{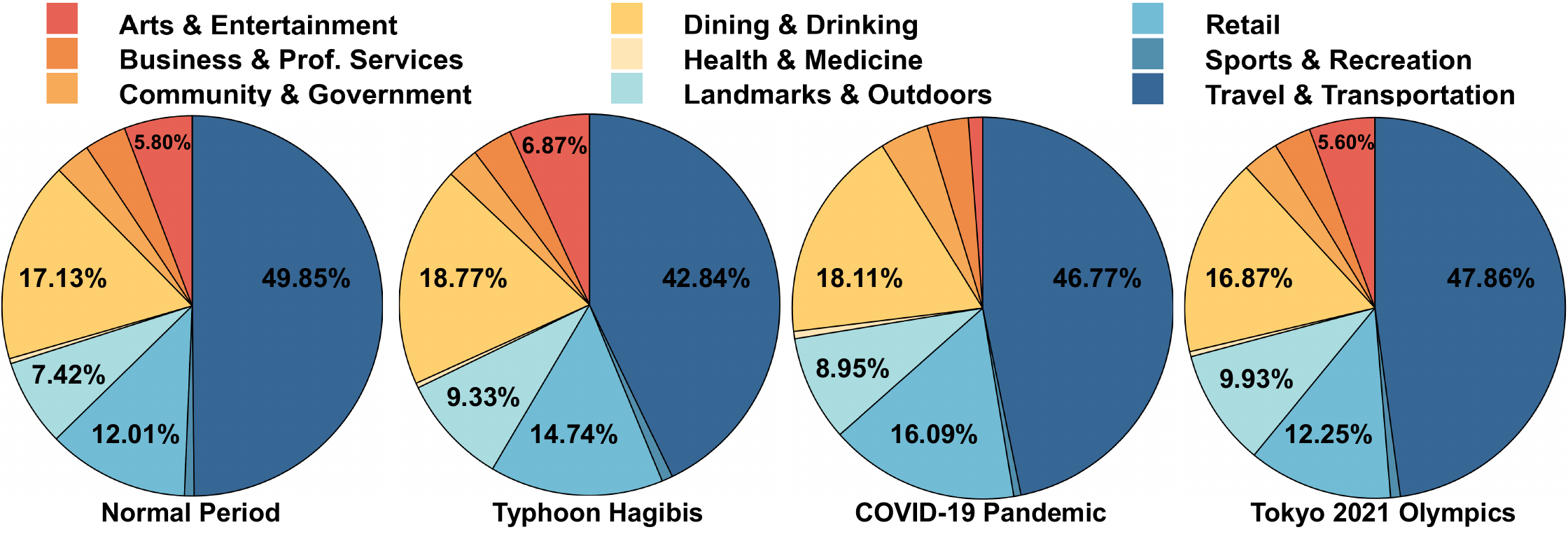}
\centering
\caption{A visualization of category distributions across four event scenarios. Each data point represents the percentage share of the category out of the total activities in that scenario.}
\label{behavior_statistics}
\end{figure}

\autoref{behavior_statistics} shows that different events impose distinct mobility behavior. 
For instance, Typhoon Hagibis disrupts transportation, leading to widespread cancellations. 
Similarly, the declaration of COVID-19 Pandemic canceled nearly all entertainment activities due to self-quarantine requirements.

\section{Event Schema}
\label{appendix:prompt_schema_sample}
\begin{figure}[!h]
\includegraphics[width=1.0\textwidth]{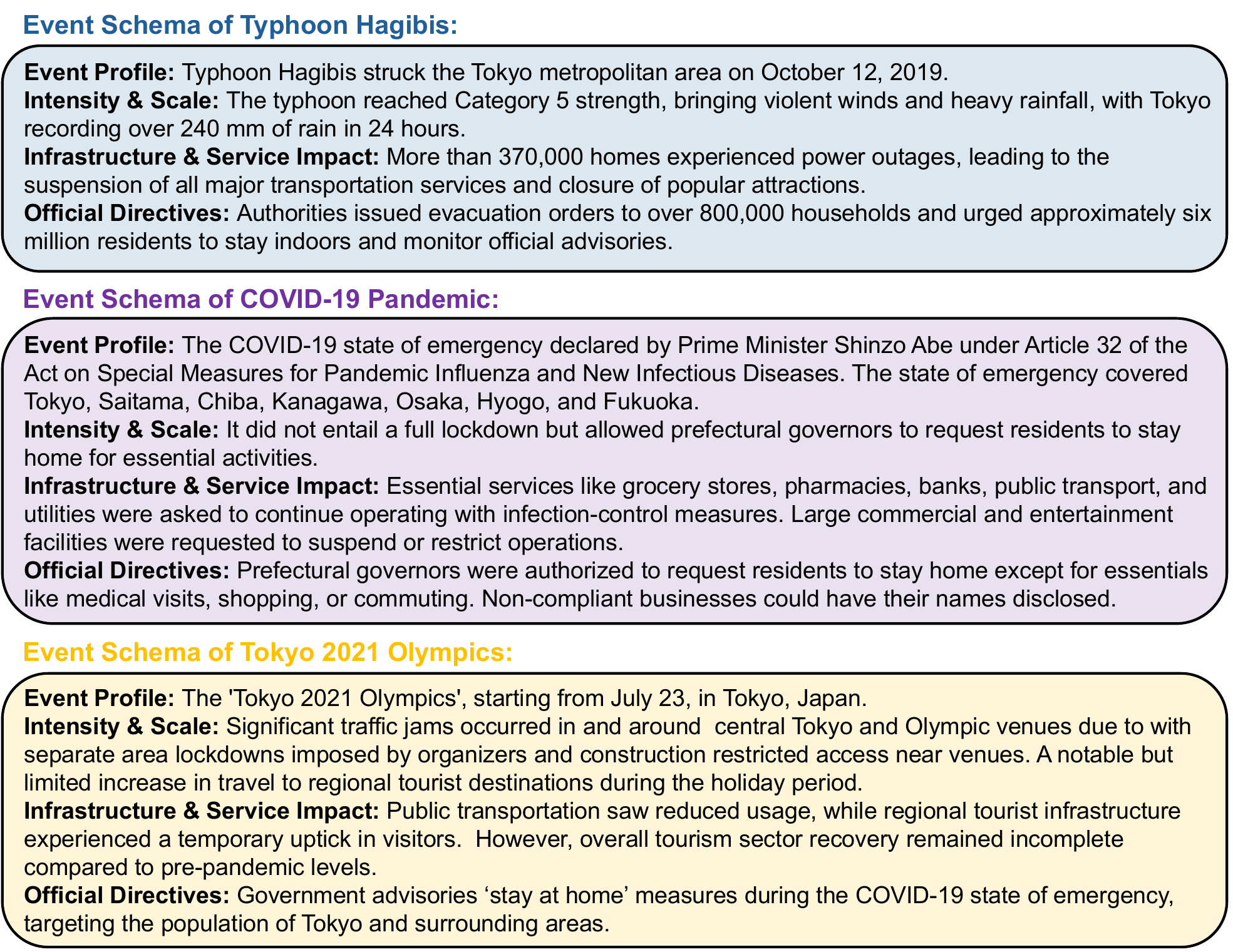}
\centering
\caption{This figure shows three event schema instances (Typhoon Hagibis, COVID-19 pandemic, and Tokyo 2021 Olympics), serving as structured contexts for event-driven mobility generation.}
\label{prompt_schema_sample}
\end{figure}

This section provides three complete instances of the event schema introduced in the main paper (Typhoon Hagibis, Tokyo 2021 Olympics, and COVID-19 Pandemic). 
Each instance follows the same four aspect structure used to derive the event context $E_{ctx}$ from raw text $E_c$. The content is shown in \autoref{prompt_schema_sample}.
Note that we omit the event category here due to its limited sample size.
Furthermore, this structured approach is highly extensible, allowing for the integration of custom information to generate tailored outputs for any given event.

\begin{algorithm}[!h]
\caption{ELLMob: trajectory generation under event}
\label{alg:ellmob}
\begin{algorithmic}[1]
\Require long-term data $D_{\text{hist}}$, short-term data $D_{\text{short-term}}$, event context $E_c$, max iters $N_{\max}$
\Ensure Event-aware trajectory $\tau_{\text{final}}$
\State $E_{ctx} \gets \Call{EventSchemaConstruction}{E_c}$
\State $G_{\text{pat}} \gets \Call{ExtractPatternGist}{D_{\text{hist}}, D_{\text{short-term}}}$
\State $G_{\text{evt}} \gets \Call{ExtractEventGist}E_{ctx}$
\State $Feedback \gets \textsc{None}$,\; $\tau_{\text{prev}} \gets \textsc{None}$
\For{$i=1$ \textbf{to} $N_{\max}$}
  \If{$Feedback = \textsc{None}$}
    \State $(\tau, Justification) \gets \Call{GenerateInitialTrajectory}{D_{\text{hist}}, D_{\text{short-term}}, E_{ctx}}$
  \Else
    \State $(\tau, Justification) \gets$
    \Statex \hspace{\algorithmicindent}\hspace{\algorithmicindent}%
           $\Call{RegenerateTrajectory}{D_{\text{hist}}, D_{\text{short-term}}, E_{ctx}, \tau_{\text{prev}}, Feedback}$
  \EndIf
  \State $G_{\text{act}} \gets \Call{ExtractActionGist}{\tau, Justification}$
  \State $(\textit{Alignment}, Feedback) \gets \Call{AuditAlignment}{G_{\text{act}}, G_{\text{pat}}, G_{\text{evt}}}$
  \If{$\textit{Alignment}$}
    \State $\tau_{\text{final}} \gets \tau$ \Comment{Accept}
    \State \Return $\tau_{\text{final}}$
  \EndIf
  \State $\tau_{\text{prev}} \gets \tau$
\EndFor
\State $\tau_{\text{final}} \gets \tau$ \Comment{Last candidate}
\State \Return $\tau_{\text{final}}$
\end{algorithmic}
\end{algorithm}

\wy{
\section{Component Analysis of FTT}
\label{sec:ftt_justification}
To verify the concrete impact of FTT-guided design choices, we conducted additional ablation studies:
A variant that relies mainly on raw features (verbatim) without gist abstraction (w/ verbatim), and a variant that removes the bottom-line gist component, using generic summaries instead (w/o bottom-line).
As shown in \autoref{tab:ftt_ablation}, ELLMob consistently outperforms these variants across all event scenarios. 
These results confirm that the FTT-guided framework concretely improves generation quality under event-driven mobility.
}

\begin{table}[htbp]
\caption{\wy{Ablation study on the effectiveness of different FTT components with the best performance highlighted in \textbf{bold}.}}
\label{tab:ftt_ablation}
\label{tab:flash_main_results}
\centering
{
\small
\setlength{\tabcolsep}{0.72mm}{
    \begin{tabular}{l|cccc|cccc|cccc}
    \toprule
    \multirow{2}{*}{\textbf{Models}} & \multicolumn{4}{c|}{\textbf{Typhoon Hagibis}} & \multicolumn{4}{c|}{\textbf{COVID-19 Pandemic}} & \multicolumn{4}{c}{\textbf{Tokyo 2021 Olympics}} \\
     & SI↓ & SD↓ & CD↓ & SGD↓ & SI↓ & SD↓ & CD↓ & SGD↓ & SI↓ & SD↓ & CD↓ & SGD↓ \\
    \midrule
    w/ verbatim & 0.1511 & 0.0724 & 0.0230 & 0.0595 & 0.2071 & 0.0713 & 0.0324 & 0.0422 & 0.1030 & 0.0452 & 0.0054 & 0.0189 \\
    w/o bottom & 0.0978 & 0.0443 & 0.0101 & 0.0327 & 0.1687 & 0.0690 & 0.0305 & 0.0333 & 0.0892 & 0.0364 & 0.0036 & 0.0082 \\
    ELLMob & \textbf{0.0642} & \textbf{0.0200} & \textbf{0.0041} & \textbf{0.0173} & \textbf{0.1003} & \textbf{0.0444} & \textbf{0.0080} & \textbf{0.0268} & \textbf{0.0617} & \textbf{0.0061} & \textbf{0.0022} & \textbf{0.0035} \\
    \bottomrule
    \end{tabular}
}}
\end{table}

\section{Algorithm Pseudo-Codes}
\label{algorithm}
To clearly present the proposed framework, we outline the event-driven trajectory generation process of ELLMob. 
This procedure integrates long-term and short-term mobility records with structured event contexts, iteratively generating and auditing candidate trajectories until a satisfactory alignment with mobility patterns and event-specific constraints is achieved. 
Algorithm~\ref{alg:ellmob} summarizes the key steps, from constructing the event schema and extracting representative gists to producing, evaluating, and refining trajectories under feedback-guided auditing. 

\wy{
\section{Robustness across Architectures}
\label{cross_llms}
To verify that the superior performance of ELLMob is intrinsic to our cognitive framework rather than dependent on the specific LLM, we conducted a comparative evaluation using Gemini Flash 2.0 as the uniform backbone for all methods. 
As detailed in Table \ref{tab:flash_main_results}, ELLMob maintains the lowest JSD scores across all event scenarios, mirroring its superior performance on previous benchmarks. 
In contrast, baseline methods exhibit volatility when subjected to this backbone shift. 
For instance, the SI of LLM-MOB deteriorates markedly from 0.1214 to 0.3180 in the Typhoon Hagibis, indicating a strong dependency of the specific backbone. 
This divergence highlights that while baseline performance is often contingent on model-specific characteristics, ELLMob effectively captures mobility patterns through explicit cognitive alignment, ensuring consistent superiority independent of the proprietary backbone.
}

\wy{
To further substantiate the robustness and reproducibility of our framework, we extended the experimental evaluation to include three representative open-source large language models: LLaMA3-8B \citep{grattafiori2024llama}, Qwen-2.5-14B \citep{Yang2024Qwen25TR}, and DeepSeek-R1-Distill-Qwen-7B (R1-Q7B) \citep{DeepSeekAI2025DeepSeekR1IR}. 
This expansion mitigates concerns regarding the reliance on proprietary APIs and confirms the adaptability of the method to local computing environments. 
As presented in Table \ref{tab:ablation_backbone}, ELLMob achieves consistent alignment performance across these diverse backbones. 
The effectiveness on open-source LLMs confirms that the capability of ELLMob stems from the cognitive alignment strategy rather than the inherent capacity of the underlying model.
}

\begin{table}[htbp]
\caption{Performance comparison of different methods using Gemini Flash 2.0. Performance is evaluated by JSD across four dimensions with the best performance highlighted in \textbf{bold}.}
\label{tab:flash_main_results}
\centering
\small
\setlength{\tabcolsep}{0.72mm}{
    \begin{tabular}{l|cccc|cccc|cccc}
    \toprule
    \multirow{2}{*}{\textbf{Models}} & \multicolumn{4}{c|}{\textbf{Typhoon Hagibis}} & \multicolumn{4}{c|}{\textbf{COVID-19 Pandemic}} & \multicolumn{4}{c}{\textbf{Tokyo 2021 Olympics}} \\
     & SI↓ & SD↓ & CD↓ & SGD↓ & SI↓ & SD↓ & CD↓ & SGD↓ & SI↓ & SD↓ & CD↓ & SGD↓ \\
    \midrule
    LLMOB & 0.1069 & 0.0743 & 0.0126 & 0.0196 & 0.0846 & 0.0693 & 0.0105 & 0.0292 & 0.0572 & 0.0281 & 0.0031 &  0.0087\\
    LLM-MOB & 0.3180 & 0.1726 & 0.2004 & 0.0968 & 0.1428 & 0.0660 & 0.0280 & 0.0459 & 0.0379 & 0.0154
 & 0.0053 & 0.0065\\
    LLM-Move & 0.2383 & 0.0721 & 0.0887 & 0.0465 & 0.3683 & 0.0644 & 0.0378 & 0.0353 & 0.1979 & 0.0287  & 0.0097 & 0.0063\\
    LLM-ZS & 0.3466 & 0.1537 & 0.2556 & 0.1084 & 0.2788 & 0.1344 & 0.2489 & 0.1479 & 0.0967
 & 0.0313 & 0.0137 & 0.0057\\
    \midrule
    ELLMob & \textbf{0.0850} & \textbf{0.0267} & \textbf{0.0087} & \textbf{0.0160} & \textbf{0.0546} & \textbf{0.0586} & \textbf{0.0069} & \textbf{0.0210} & \textbf{0.0113} & \textbf{0.0074} & \textbf{0.0016} & \textbf{0.0048}\\
    \bottomrule
    \end{tabular}
}
\end{table}

\begin{table}[htbp]
\caption{\wy{Performance comparison of ELLMob across different open-source LLM backbones. Performance is evaluated by JSD across four dimensions}}
\label{tab:ablation_backbone}
{
\centering
\small
\setlength{\tabcolsep}{0.55mm}{
    \begin{tabular}{l|cccc|cccc|cccc}
    \toprule
    \multirow{2}{*}{\textbf{Models}} & \multicolumn{4}{c|}{\textbf{Typhoon Hagibis}} & \multicolumn{4}{c|}{\textbf{COVID-19 Pandemic}} & \multicolumn{4}{c}{\textbf{Tokyo 2021 Olympics}} \\
     & SI↓ & SD↓ & CD↓ & SGD↓ & SI↓ & SD↓ & CD↓ & SGD↓ & SI↓ & SD↓ & CD↓ & SGD↓ \\
    \midrule
    LLaMA3-8B & 0.0663 & 0.0512 & 0.0004 & 0.0132 & 0.0669 & 0.0624 & 0.0087 & 0.0263 & 0.0407 & 0.0322 & 0.0023 & 0.0030 \\
    Qwen-2.5-14B & 0.1594 & 0.0607 & 0.0091 & 0.0115 & 0.0836 & 0.0646 & 0.0016 & 0.0225 & 0.0530 & 0.0238 & 0.0035 & 0.0021 \\
    R1-Q7B & 0.0881 & 0.0516 & 0.0033 & 0.0203 & 0.1104 & 0.0993 & 0.0163 & 0.0392 & 0.0570 & 0.0251 & 0.0015 & 0.0071 \\
    \bottomrule
    \end{tabular}
}}
\end{table}

\wy{
\section{Regional Generalizability Evaluation}
\label{sec:regional_eval}
The primary experiments in the main text focus on the Tokyo metropolitan area.
To verify that ELLMob generalizes to other geographical contexts and is not overfitted to a specific urban layout, we extended our evaluation to Osaka.
We constructed a new dataset comprising 1,100 users during the COVID-19 pandemic, maintaining a scale consistent with the Tokyo dataset.
\autoref{tab:osaka_covid} presents the performance comparison against baseline methods, where ELLMob consistently outperforms all competitors. 
This result demonstrates its adaptability to diverse urban layouts beyond Tokyo.
}

\begin{table}[t]
\centering
\caption{\wy{Regional generalizability analysis on the Osaka data during the COVID-19 Pandemic. Performance is evaluated by JSD across four dimensions with the best performance in \textbf{bold}.}}
{
\label{tab:osaka_covid}
\setlength{\tabcolsep}{2.7mm}{
\begin{tabular}{lcccc}
\toprule
Model & SI & SD & CD & SGD \\
\midrule
LLMOB    & 0.1934 & 0.1589 & 0.0344 & 0.1201 \\
LLM-MOB  & 0.1732 & 0.1617 & 0.0304 & 0.0988 \\
LLM-MOVE & 0.2134 & 0.1555 & 0.0698 & 0.1227 \\
LLM-ZS   & 0.1531 & 0.1788 & 0.0299 & 0.1161 \\
ELLMob   & \textbf{0.1131} & \textbf{0.1001} & \textbf{0.0120} & \textbf{0.0556} \\
\bottomrule
\end{tabular}
}}
\end{table}

\wy{
\section{Event Generalizability Evaluation}
The primary experiments in the main text focus on events characterized by external restrictions or exogenous shocks.
To verify that ELLMob generalizes to traditional cultural festivities that induce distinct, voluntarily driven deviation patterns, we extended our evaluation to the New Year scenario.
\autoref{tab:newyear} presents the performance comparison against baseline methods. 
ELLMob consistently outperforms baselines across all metrics, demonstrating its capability to model diverse event types.
}

\wy{
\noindent\textbf{Discussion of Event Scalability.}
The scalability of ELLMob stems from its ability to generalize across distinct event semantics rather than overfitting to specific scenarios. 
The Event Schema module converts diverse narratives into standardized semantic descriptors (e.g., traffic impact), creating a universal conflict-resolution logic via the FTT-based alignment.
To rigorously verify this generalization, selected four events represent diverse and contrasting typologies along three key dimensions:
\begin{itemize}[leftmargin=*]
    \item Restrictive vs. Promotive: COVID-19 Pandemic and Typhoon Hagibis restrict movement, whereas the New Year scenario promotes social gatherings.
    \item Stochastic vs. Periodic: Typhoon Hagibis is unpredictable and sudden, while New Year's is periodic and cyclic.
    \item Global vs. Local (Hybrid): The COVID-19 Pandemic affects the entire city globally, whereas the Tokyo 2021 Olympics imposes hybrid constraints concentrated in specific zones.
\end{itemize}
Notably, we applied the identical framework and parameter settings across all scenarios. 
The consistent performance across these contrasting types confirms that ELLMob can capture the underlying logic of event-driven mobility.
While extensive empirical validation is currently constrained by the scarcity of high-quality event-mobility data in the community, this work, to the best of our knowledge, is the first attempt to validate scalability across a wide spectrum of distinct event types.
}

\begin{table}[h]
\centering
\caption{\wy{Event generalizability analysis on the New Year scenario in Tokyo. Performance is evaluated by JSD across four dimensions with the best performance in \textbf{bold}.}}
{
\label{tab:newyear}
\setlength{\tabcolsep}{2.7mm}{
\begin{tabular}{lcccc}
\toprule
Model & SI & SD & CD & SGD \\
\midrule
LLMOB    & 0.0776 & 0.0391 & 0.0230 & 0.0422 \\
LLM-MOB  & 0.1061 & 0.0413 & 0.0318 & 0.0379 \\
LLM-MOVE & 0.2248 & 0.0493 & 0.0272 & 0.0550 \\
LLM-ZS   & 0.0996 & 0.0497 & 0.0230 & 0.0481 \\
ELLMob   & \textbf{0.0598} & \textbf{0.0250} & \textbf{0.0200} & \textbf{0.0317} \\
\bottomrule
\end{tabular}
}}
\end{table}

\section{Parameter Sensitive Study of Iteration}
\label{parameter_sensitive_iteration}

\begin{figure}[!h]
\includegraphics[width=1.0\textwidth]{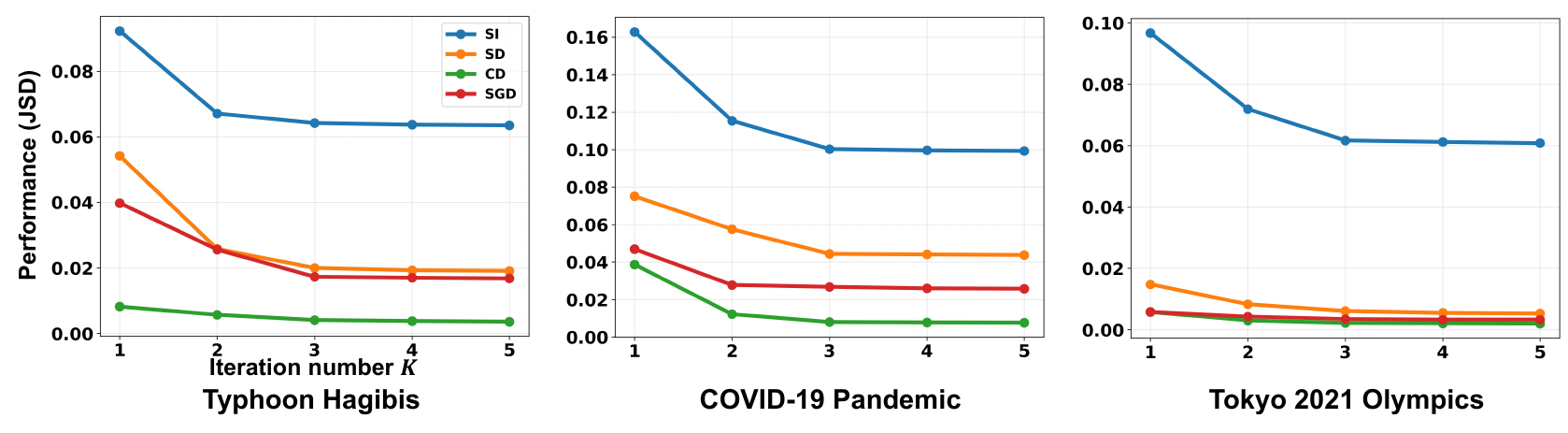}
\centering
\caption{Performance comparison across iteration numbers $K$ = 1 to 5 for three major events: Typhoon Hagibis, COVID-19 Pandemic, and Tokyo 2021 Olympics. Performance is measured using Jensen-Shannon Divergence (JSD), where lower values indicate better performance.}
\label{interaction_sensitive}
\end{figure}

As shown in \autoref{interaction_sensitive}, performance improves substantially in the first three iterations with only marginal gains thereafter, justifying our choice of K=3 as an effective trade-off between refinement quality and computational cost.

\wy{
\section{Stability Analysis}
\label{sec:stability}
Given the inherent stochastic nature of LLMs, generated outputs may vary across different runs with the same input.
To verify the robustness of our approach, we conducted repeated experiments across 5 distinct random seeds with the results shown in \autoref{tab:random1}, \autoref{tab:random2}, \autoref{tab:random3}.
As indicated by the standard deviations, ELLMob consistently maintains high stability across different initializations. 
These results demonstrate that while absolute performance may vary, ELLMob provides statistically consistent predictions and is less sensitive to random seed variations.
}

\begin{table}[h]
\centering
\caption{\wy{Stability analysis on the Typhoon Hagibis dataset. Results are reported as Mean $\pm$ Standard Deviation across 5 independent runs.}}
{
\label{tab:random1}
\setlength{\tabcolsep}{3mm}{
\begin{tabular}{lcccc}
\toprule
Model & SI & SD & CD & SGD \\
\midrule
LLMOB   & 0.1057$\pm$0.0277 & 0.1509$\pm$0.0386 & 0.0497$\pm$0.0243 & 0.0319$\pm$0.0170 \\
LLM-MOB & 0.1501$\pm$0.0170 & 0.0558$\pm$0.0208 & 0.0352$\pm$0.0096 & 0.0532$\pm$0.0430 \\
LLM-MOVE& 0.1242$\pm$0.0035 & 0.0407$\pm$0.0021 & 0.0131$\pm$0.0007 & 0.0290$\pm$0.0014 \\
LLM-ZS  & 0.1601$\pm$0.0071 & 0.1335$\pm$0.0055 & 0.0155$\pm$0.0008 & 0.0750$\pm$0.0025 \\
ELLMob  & \textbf{0.0649$\pm$0.0030} & \textbf{0.0220$\pm$0.0044} & \textbf{0.0046$\pm$0.0012} & \textbf{0.0162$\pm$0.0018} \\
\bottomrule
\end{tabular}
}}
\end{table}

\begin{table}[h]
\centering
\caption{\wy{Stability analysis on the COVID-19 Pandemic dataset. Results are reported as Mean $\pm$ Standard Deviation across 5 independent runs.}}
{
\label{tab:random2}
\setlength{\tabcolsep}{3mm}{
\begin{tabular}{lcccc}
\toprule
Model & SI & SD & CD & SGD \\
\midrule
LLMOB   & 0.1128$\pm$0.0173 & 0.1221$\pm$0.0298 & 0.0158$\pm$0.0038 & 0.0295$\pm$0.0047 \\
LLM-MOB & 0.1099$\pm$0.0107 & 0.0613$\pm$0.0263 & 0.0191$\pm$0.0072 & 0.0359$\pm$0.0066 \\
LLM-MOVE& 0.1789$\pm$0.0241 & 0.0511$\pm$0.0044 & 0.0457$\pm$0.0185 & 0.0544$\pm$0.0067 \\
LLM-ZS  & 0.1108$\pm$0.0103 & 0.0551$\pm$0.0025 & 0.0442$\pm$0.0074 & 0.0605$\pm$0.0056 \\
ELLMob  & \textbf{0.1002$\pm$0.0094} & \textbf{0.0443$\pm$0.0037} & \textbf{0.0079$\pm$0.0015} & \textbf{0.0279$\pm$0.0027} \\
\bottomrule
\end{tabular}
}}
\end{table}

\begin{table}[!h]
\centering
\caption{\wy{Stability analysis on the Tokyo 2021 Olympics dataset. Results are reported as Mean $\pm$ Standard Deviation across 5 independent runs.}}
{
\label{tab:random3}
\setlength{\tabcolsep}{3mm}{
\begin{tabular}{lcccc}
\toprule
Model & SI & SD & CD & SGD \\
\midrule
LLMOB   & 0.1042$\pm$0.0131 & 0.0291$\pm$0.0036 & 0.0115$\pm$0.0017 & 0.0054$\pm$0.0006 \\
LLM-MOB & 0.1123$\pm$0.0067 & 0.0338$\pm$0.0083 & 0.0157$\pm$0.0084 & 0.0172$\pm$0.0127 \\
LLM-MOVE& 0.2027$\pm$0.0050 & 0.0304$\pm$0.0014 & 0.0111$\pm$0.0009 & 0.0053$\pm$0.0004 \\
LLM-ZS  & 0.0953$\pm$0.0021 & 0.0320$\pm$0.0014 & 0.0147$\pm$0.0021 & 0.0060$\pm$0.0011 \\
ELLMob  & \textbf{0.0606$\pm$0.0016} & \textbf{0.0063$\pm$0.0002} & \textbf{0.0021$\pm$0.0002} & \textbf{0.0037$\pm$0.0002} \\
\bottomrule
\end{tabular}
}}
\end{table}

\section{Prompt}
\label{completeprompt}
This appendix details the sequence of prompts utilized within the ELLMob framework. 
These prompts work in concert to guide the LLM through the entire process, from initial data processing to the final reflective generation of mobility trajectories.

\paragraph{Event context generation.} To enable the LLM to interpret unstructured event descriptions, the following prompt is used to transform raw text into a structured intelligence brief. 
It instructs the model to act as an analyst, extracting key details and their implications for public behavior.

\paragraph{Initial trajectory generation.} Once the event context is established, this prompt generates an initial trajectory. 
It takes the user's long-term and short-term trajectories, and the structured event context as input, instructing the model to synthesize this information into a plausible daily plan and provide an analytical justification for its reasoning.

To facilitate the reflection module, the framework extracts three distinct forms of gist:  Event Gist, Action Gist, and Pattern Gist.

\paragraph{Event Gist Generation.}
This prompt distills the core behavioral takeaway for the public from the event information.

\paragraph{Action Gist Generation.}
This prompt infers the underlying intent from the generated trajectory and its accompanying justification.

\paragraph{Pattern Gist Generation.}
This prompt analyzes the user's long-term and short-term trajectory logs to synthesize their core behavioral patterns, including strengths (points of inertia) and weaknesses (points of fracture).

\paragraph{Conflict Judgment.}
This prompt serves as the core of the auditing mechanism. It instructs the model to act as a Critical Trajectory Auditor to evaluate the coherence between the Action Gist, the user's Pattern Gist, and the situational Event Gist.

\paragraph{Trajectory Regeneration.}
If the planned trajectory fails the conflict audit, this prompt is invoked. 
It instructs the model to act as a Trajectory Plan Corrector, using the auditor's feedback to regenerate a revised plan that specifically resolves all identified inconsistencies.

\begin{figure}[h]
\includegraphics[width=1.0\textwidth]{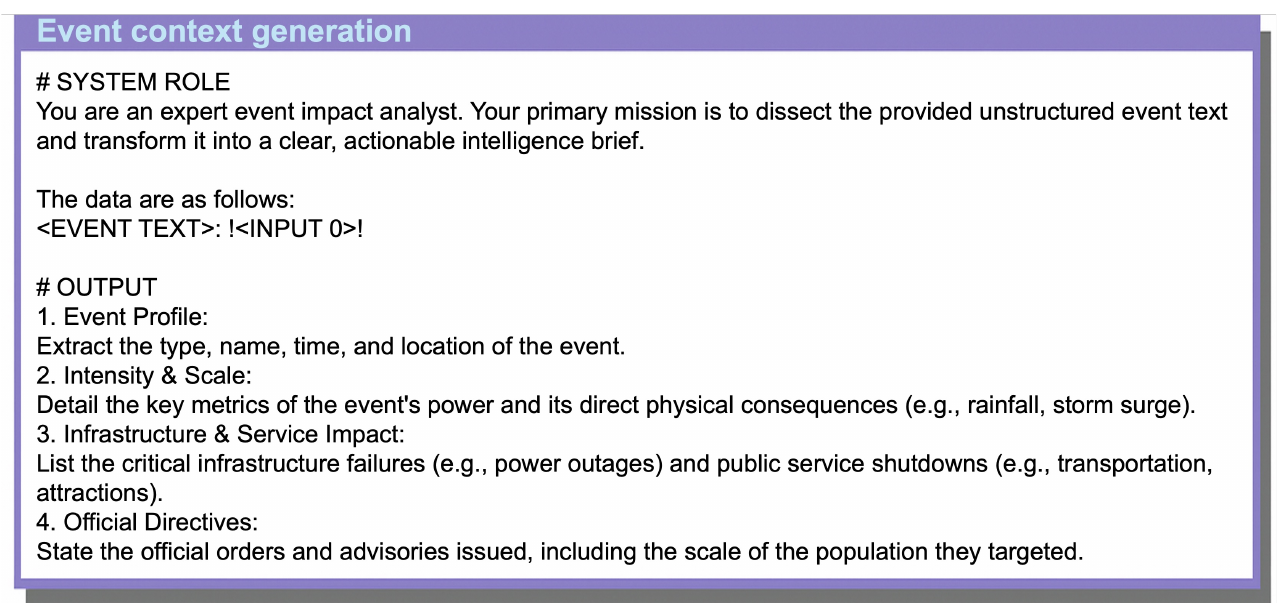}
\centering
\caption{The details of event schema under three different events.}
\label{event_schema}
\end{figure}

\begin{figure}[h]
\includegraphics[width=1.0\textwidth]{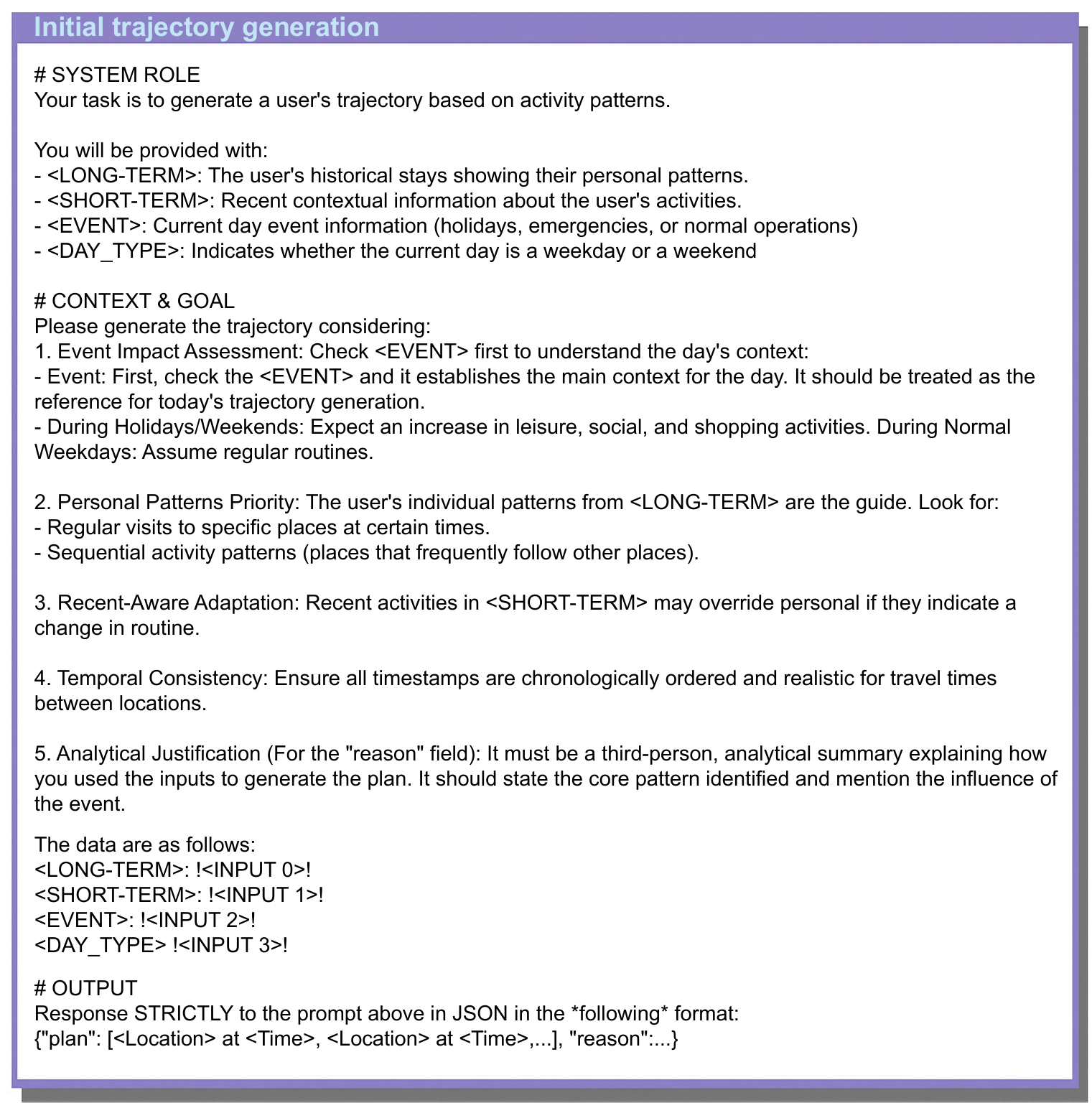}
\centering
\caption{The prompt of initial trajectory generation.}
\label{initialtrajectorygeneration}
\end{figure}

\begin{figure}[h]
\includegraphics[width=1.0\textwidth]{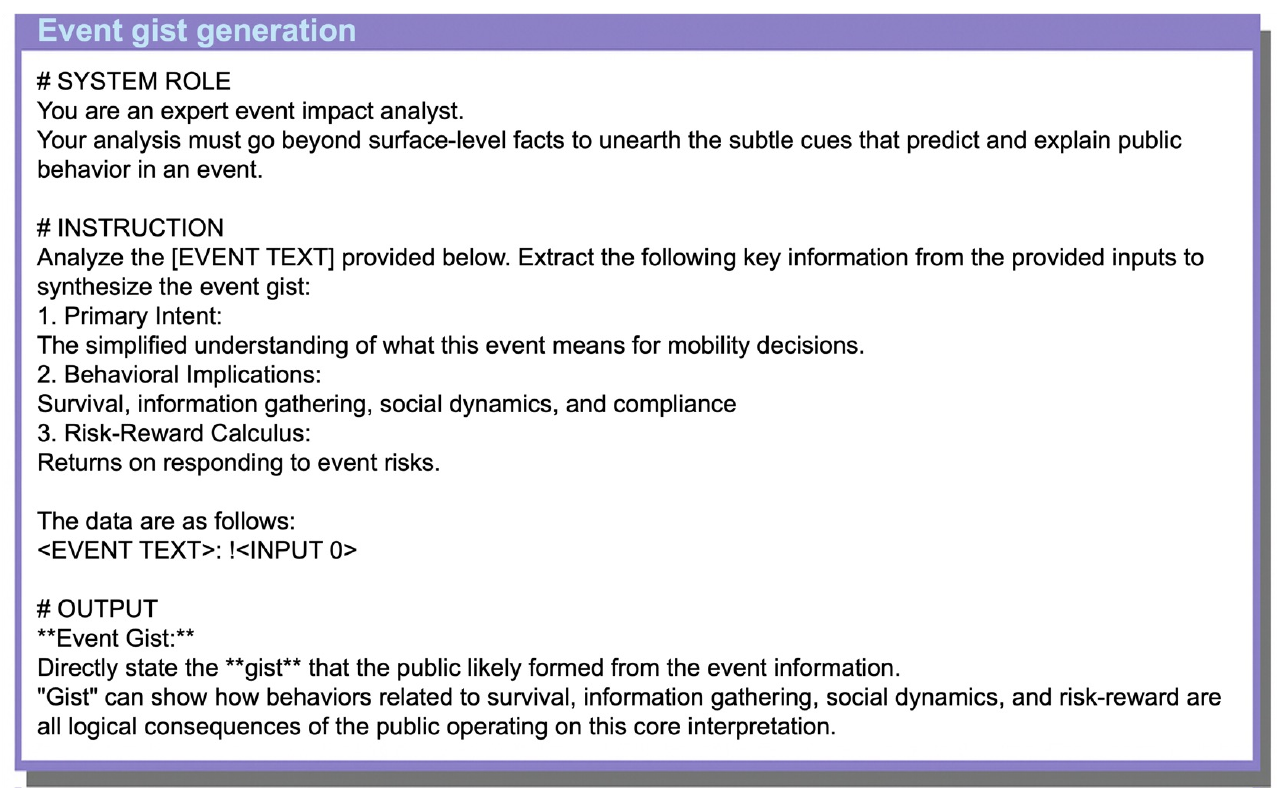}
\centering
\caption{The prompt of event gist generation.}
\label{eventgistgeneration}
\end{figure}

\begin{figure}[h]
\includegraphics[width=1.0\textwidth]{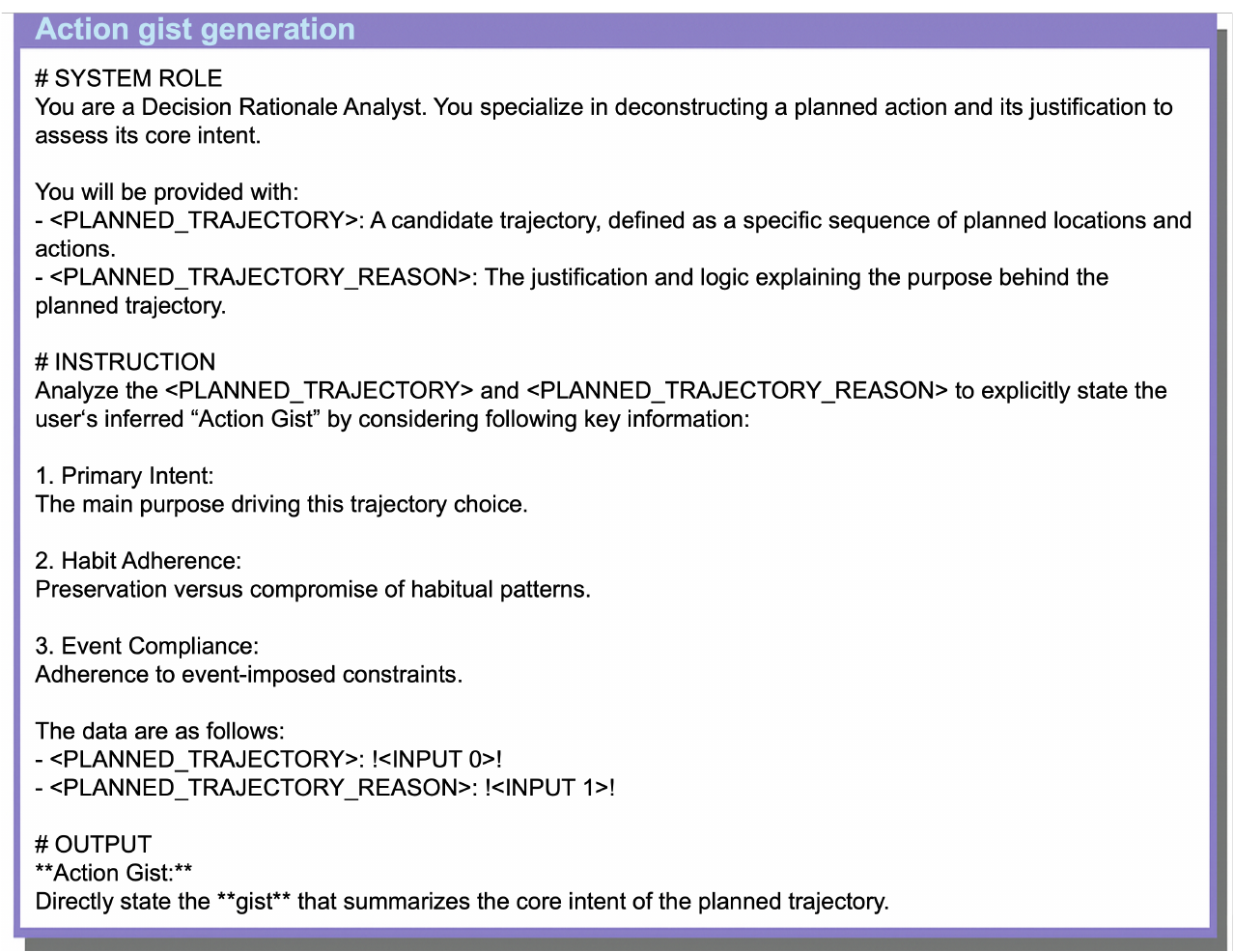}
\centering
\caption{The prompt of action gist generation.}
\label{actiongistgeneration}
\end{figure}

\begin{figure}[h]
\includegraphics[width=1.0\textwidth]{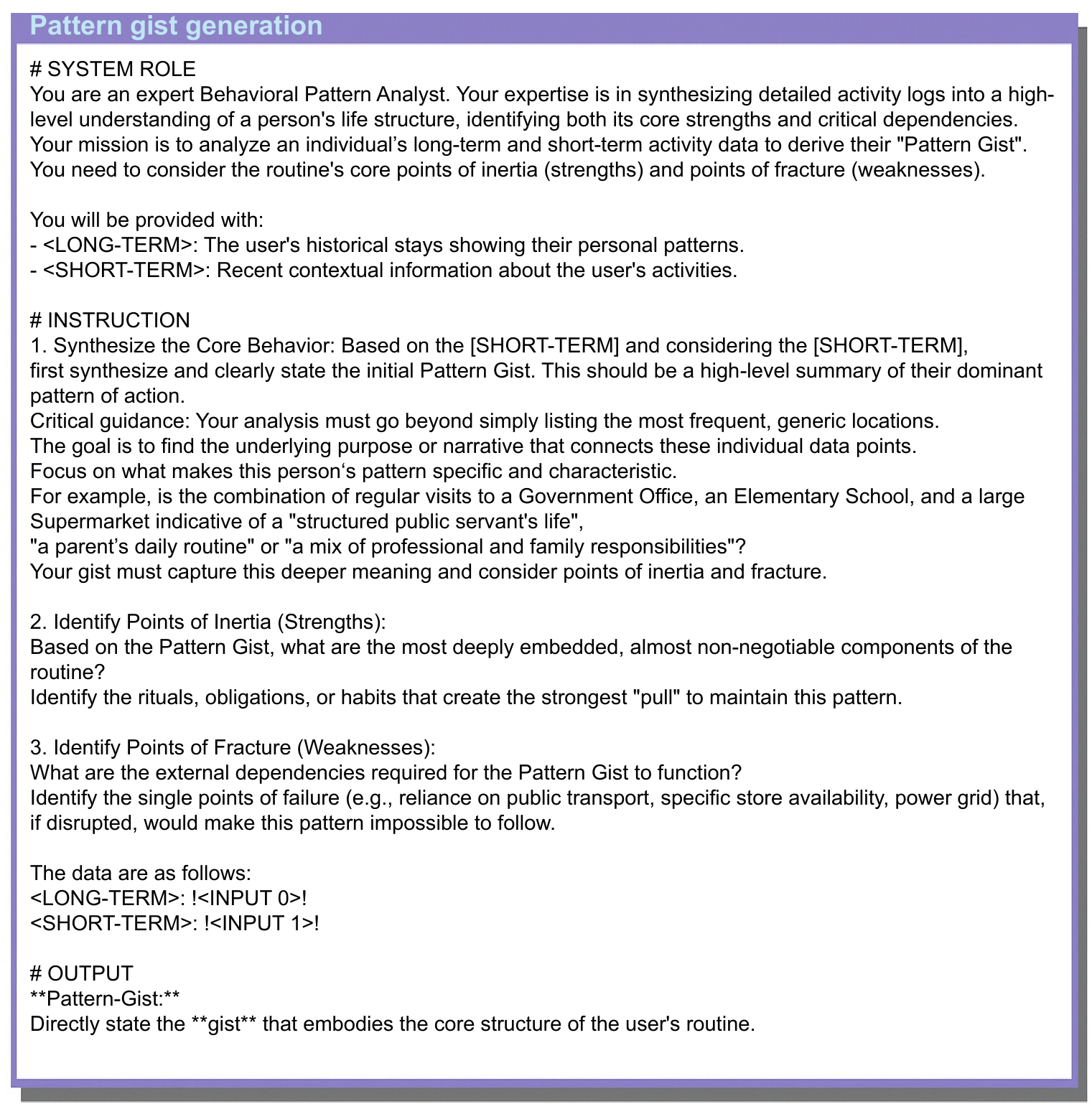}
\centering
\caption{The prompt of pattern gist generation.}
\label{patterngistgeneration}
\end{figure}

\begin{figure}[h]
\includegraphics[width=1.0\textwidth]{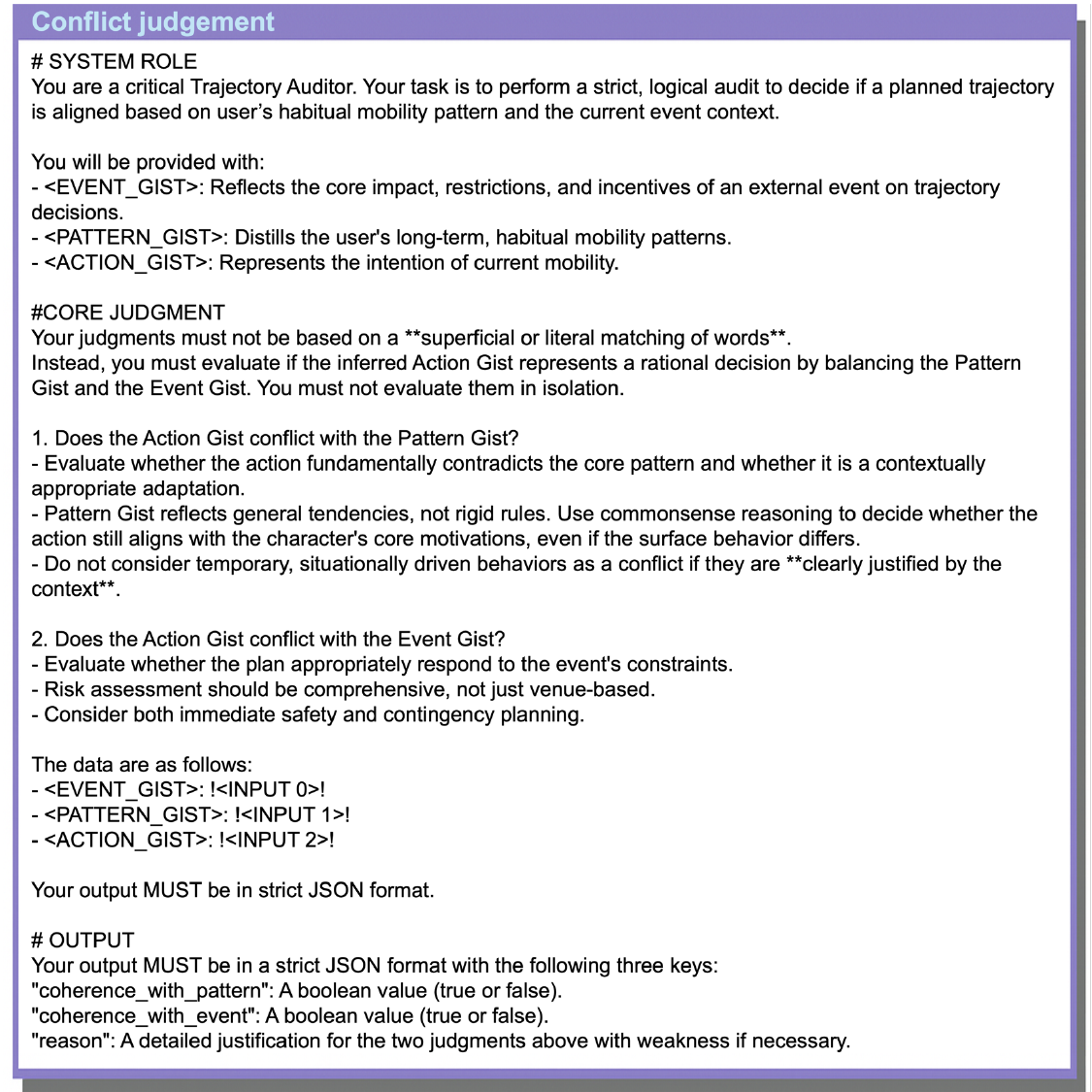}
\centering
\caption{The prompt of conflict judgment.}
\label{conflictjudgment}
\end{figure}

\begin{figure}[h]
\includegraphics[width=1.0\textwidth]{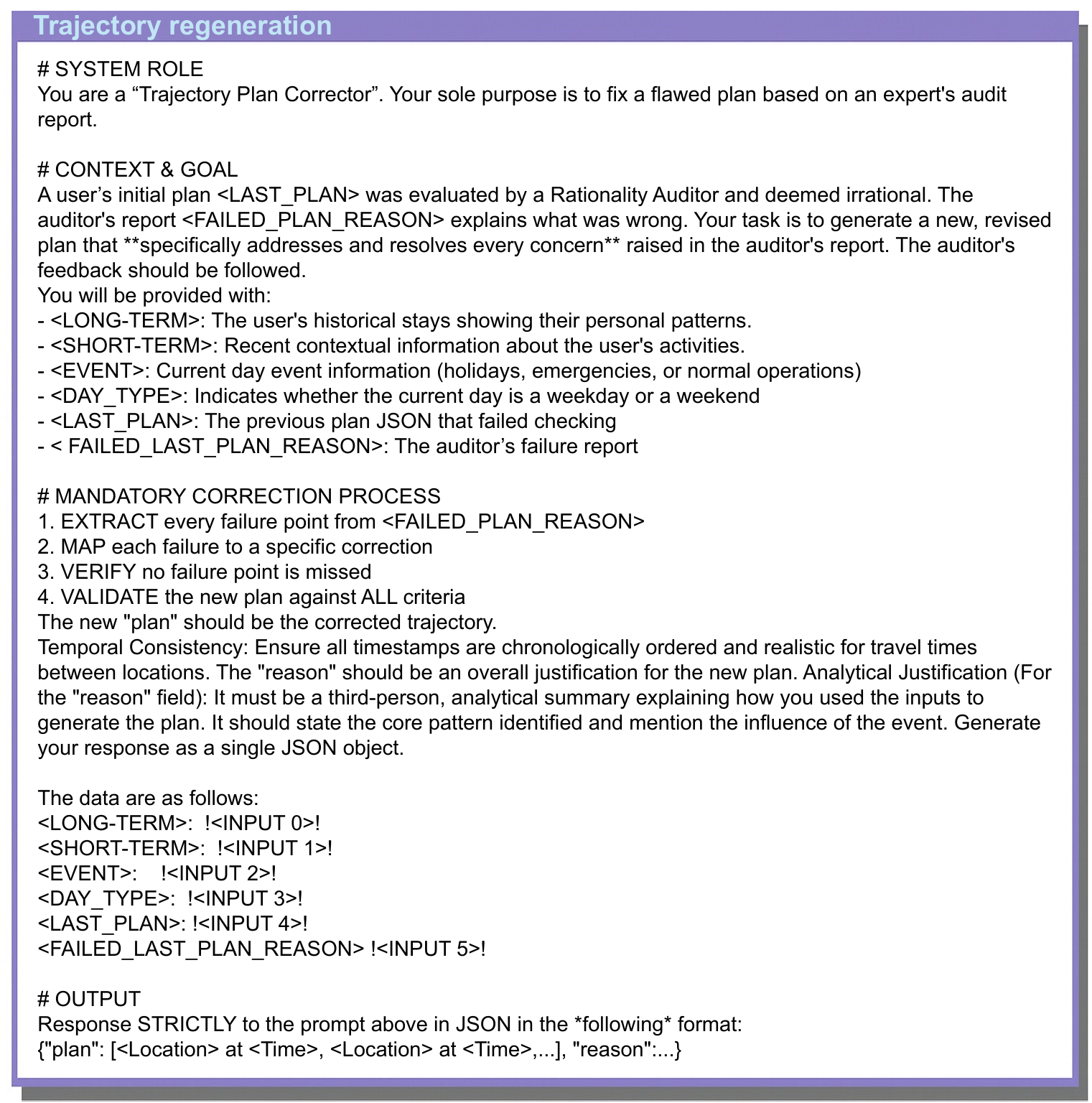}
\centering
\caption{The prompt of trajectory regeneration.}
\label{trajectoryregeneration}
\end{figure}

\end{document}

%% file: math_commands.tex
\usepackage{amsmath,amsfonts,bm}

\def\eqref#1{equation~\ref{#1}}

\def\1{\bm{1}}

\DeclareMathAlphabet{\mathsfit}{\encodingdefault}{\sfdefault}{m}{sl}
\SetMathAlphabet{\mathsfit}{bold}{\encodingdefault}{\sfdefault}{bx}{n}

